\documentclass[10pt,letterpaper]{article}
\usepackage[pagenumbers]{cvpr} 

\definecolor{cvprblue}{rgb}{0.21,0.49,0.74}
\usepackage[pagebackref,breaklinks,colorlinks,allcolors=cvprblue]{hyperref}
\usepackage{tikz}
\usetikzlibrary{positioning,calc,arrows.meta}
\def\paperID{16941} 
\def\confName{CVPR}
\def\confYear{2026}

\title{Do vision models perceive illusory motion in static images like humans?}
\author{Isabella E. Rosario*$^1$, Fan L. Cheng*$^{\dagger 1}$, Zitang Sun$^2$, Nikolaus Kriegeskorte$^1$\\
$^1$Columbia University \quad $^2$Kyoto University
}

\begin{document}
\maketitle
\renewcommand{\thefootnote}{\fnsymbol{footnote}}
\footnotetext[1]{These authors contributed equally to this work.}
\footnotetext[2]{Corresponding author: fan.cheng@columbia.edu.}

\begin{abstract}

\noindent Understanding human motion processing is essential for building reliable, human-centered computer vision systems. Although deep neural networks (DNNs) achieve strong performance in optical flow estimation, they remain less robust than humans and rely on fundamentally different computational strategies. Visual motion illusions provide a powerful probe into these mechanisms, revealing how human and machine vision align or diverge. While recent DNN-based motion models can reproduce dynamic illusions such as reverse-phi, it remains unclear whether they can perceive illusory motion in static images, exemplified by the Rotating Snakes illusion. We evaluate several representative optical flow models on Rotating Snakes and show that most fail to generate flow fields consistent with human perception. Under simulated conditions mimicking saccadic eye movements, only the human-inspired Dual-Channel model exhibits the expected rotational motion, with the closest correspondence emerging during the saccade simulation. Ablation analyses further reveal that both luminance-based and higher-order color--feature--based motion signals contribute to this behavior and that a recurrent attention mechanism is critical for integrating local cues. Our results highlight a substantial gap between current optical-flow models and human visual motion processing, and offer insights for developing future motion-estimation systems with improved correspondence to human perception and human-centric AI.
\end{abstract}

\section{Introduction}

Interpretation of visual motion information is crucial for successfully navigating and interacting with complex, dynamic environments, including those encountered in autonomous driving, object tracking, robotics, and video interpolation. Humans possess a sophisticated visual system capable of performing diverse tasks with remarkable stability and robust generalization to new environments, while consuming little energy.

Although state-of-the-art computer vision models now surpass human performance on many motion estimation benchmarks~\cite{yang2023psychophysical}, they exhibit well-documented limitations. For example, deep neural networks (DNNs) for
optical flow often struggle with occlusions, large displacements, and non-rigid motion, and can behave unpredictably under experimental conditions that humans handle robustly~\cite{ren2017unsupervised, dosovitskiy2015flownet, chen2022motion, ranjan2019attacking}. These discrepancies suggest that the human visual system relies on computational strategies not captured by models that track per-pixel displacements alone. A deeper understanding of the principles underlying human motion perception may therefore inspire more robust and efficient machine vision systems.

One promising direction is to evaluate models through the lens of human perception. Illusory motion phenomena provide a powerful testbed for adjudicating among computational models~\cite{weiss2002motion,watanabe2018illusory,kobayashi2022motion,kirubeswaran2023inconsistent}. 
From both engineering and scientific perspectives, models that reproduce human perceptual biases are valuable: they may generalize better in real-world settings and offer mechanistic insight into the computations of the visual system~\cite{yang2025illusions}. Recent neuroscience-informed motion models incorporating motion-energy filtering, higher-order pathways, and recurrent integration~\cite{meattention,tangemann2024object,sun2025machine} have successfully replicated classic motion illusions such as reverse-phi and the barber-pole effect, suggesting potential shared computational principles with biological vision. However, it remains unclear whether modern DNNs can reproduce \emph{anomalous-motion 
illusions} such as the Rotating Snakes~\cite{kitaoka_phenomenal_2003,Otero-Millan6043}, whose perceptual dynamics depend on subtle luminance asymmetries and fixational eye movements~\cite{lotter2016deep, watanabe2018illusory, kobayashi2022motion, kirubeswaran2023inconsistent}.

This work addresses this gap using an \emph{in silico} psychophysical approach to systematically probe the correspondence between deep models and human motion perception. Our contributions are summarized below:

\begin{itemize}
    \item We comprehensively evaluate a broad set of representative DNN-based and human-inspired motion models to assess their ability to reproduce \emph{illusory motion in static images}, with a primary
    focus on the \emph{Rotating Snakes}.

    \item We propose a unified experimental pipeline and validation benchmark that integrates controlled comparisons, simulated viewing conditions, and quantitative evaluation metrics. Under specific conditions, some neuroscience-inspired models exhibit human-like motion percepts that are not reproduced by conventional DNNs.

    \item We perform systematic {probing, control and ablation analyses} to identify the computational components responsible for illusion replication. Our results highlight key mechanisms, such as dual-channel motion processing, oculomotor
    transients, and recurrent integration, may support more human-like, robust generalization in complex visual environments.
\end{itemize}
\section{Experimental Design and Method}
We analyzed anomalous motion illusions (i.e., illusory motion in static images), focusing on the Rotating Snakes illusion \cite{kitaoka_phenomenal_2003}. 

\subsection{Image Design}

\subsubsection{Rotating Snakes Illusion}

The Rotating Snakes variant is among the most extensively characterized anomalous motion illusions in human psychophysics and primate neurophysiology ~\cite{kitaoka_phenomenal_2003, conway_neural_2005, atala-gerard_rotating_2017, bach_rotating_2020, fermuller_illusory_2010, kitaoka_color-dependent_2014, murakami_positive_2006, mather2025pupil, Otero-Millan6043, uesaki2024blue, ashida2012direction}. It induces highly predictable direction of perceived rotation for human observers, and its parametric stimulus construction affords precise experimental control, allowing rigorously matched illusion and control sets and clean parametric sweeps.  

\begin{figure}[b]
  \centering
  \includegraphics[width=0.5\linewidth,height=0.5\textheight,keepaspectratio]{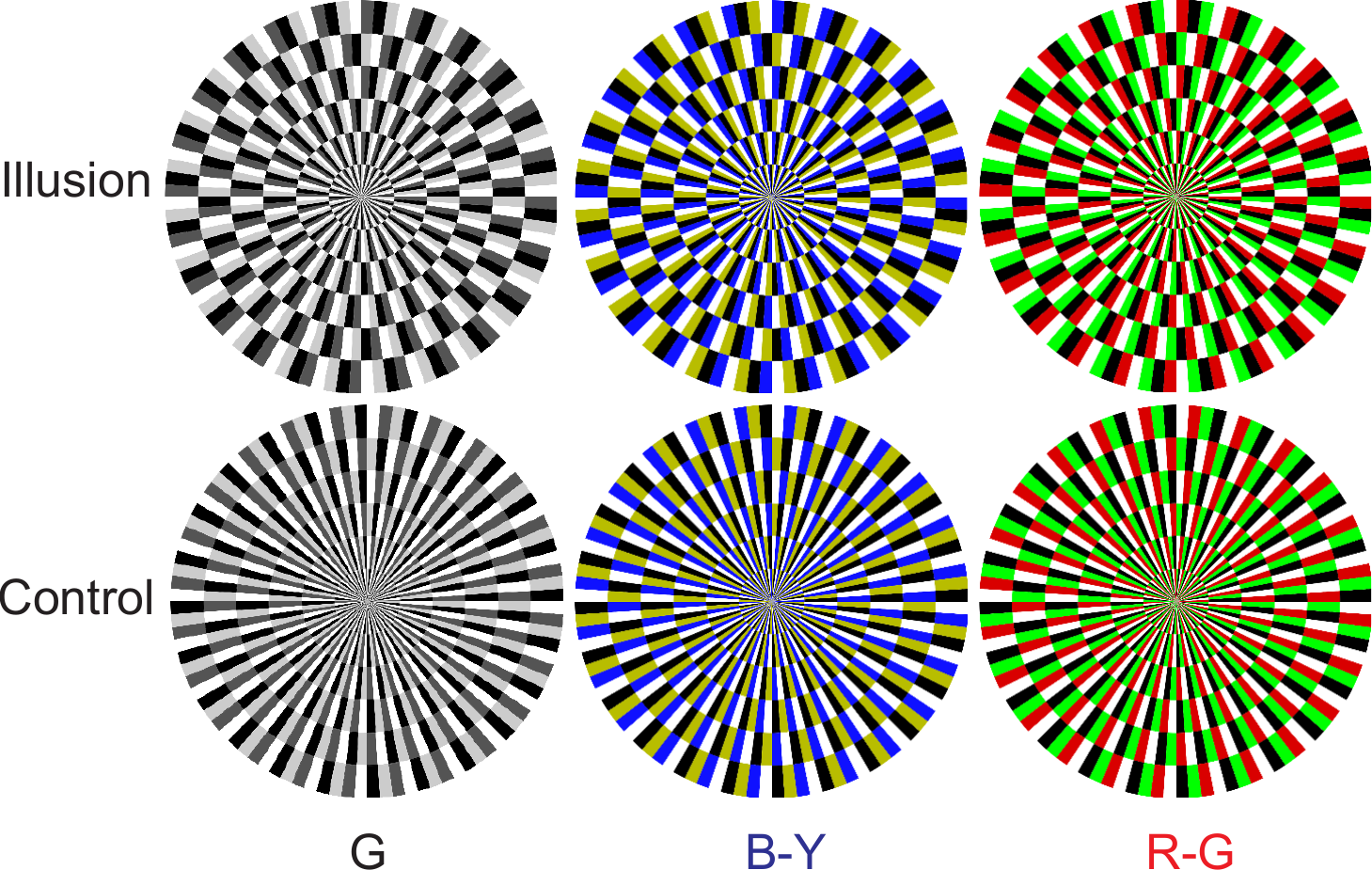}
  \caption{\textbf{Rotating Snakes illusory and control images.} 
Top row: illusory images in grayscale (G), blue–yellow (B-Y), and red–green (R-G) color schemes, each producing a robust perception of \emph{counterclockwise} rotation in human observers. 
Bottom row: corresponding control images created by permuting the luminance or color sequence within each repeating unit, which fails to induce perception of motion while preserving global layout.}
  \label{fig:fig1}
\end{figure}

We generated Rotating Snakes illusion and control images (Figure~\ref{fig:fig1}) following established stimulus construction procedures~\cite{uesaki2024blue}. Each illusory image consisted of six concentric rings, each containing 24 repeating elements. Images were rendered using either grayscale (G), blue--yellow (B-Y), or red--green (R-G) color schemes, all of which have been shown to reliably induce counterclockwise rotation for most human observers. Control images were created by permuting the luminance or color sequence within each repeating unit, thereby eliminating the local luminance gradients responsible for driving the illusion \cite{atala-gerard_rotating_2017}. These controls preserved the overall spatial layout and chromatic composition while disrupting the specific asymmetric micro-patterns known to generate perceived rotation. 

\subsubsection{Other Anomalous Motion illusions}
\label{sec:other_anom_motion}

Beyond the Rotating Snakes illusion, we also analyzed a broader set of anomalous motion illusions. These patterns are grouped according to Kitaoka's tentative classification of anomalous motion illusions~\cite{kitaoka_classify}, which organizes them by the image features that drive illusory motion. Figure~\ref{fig:fig6} show the specific exemplars used in our experiments.

Peripheral drift illusions correspond to a broader family that includes the Rotating Snakes illusions~\cite{faubert1999peripheral}. Observers report that the rings rotate or “creep” around the center when they move their eyes, blink, or experience a sudden stimulus onset. The exemplars sample several geometries (\eg, nested circular arrangements, radial spokes, and logarithmic spirals) but share the same core design principle: a repeating sequence of unequal luminance steps that biases the motion system under transient retinal slips.

In central drift illusions~\cite{kitaoka_central_drift}, radial-shaped sectors appear to rotate even when viewed centrally, with perceived motion aligned with the underlying luminance gradient. 

Ouchi illusions are characterized by apparent relative motion between regions of different orientation or spatial frequency during retinal slips ~\cite{hine1997ouchi}. A central patch and its surround are composed of orthogonal line or check arrangements. During slight image motion or eye movements, the central region appears to drift or jitter relative to the background, often in a direction roughly perpendicular to the physical motion of the stimulus. 

\subsection{Model Suite and Evaluation Settings}

\subsubsection{Pretrained Models and Inference Details}

\begin{figure*}[b]
  \centering
  \includegraphics[width=\linewidth,height=0.72\textheight,keepaspectratio]{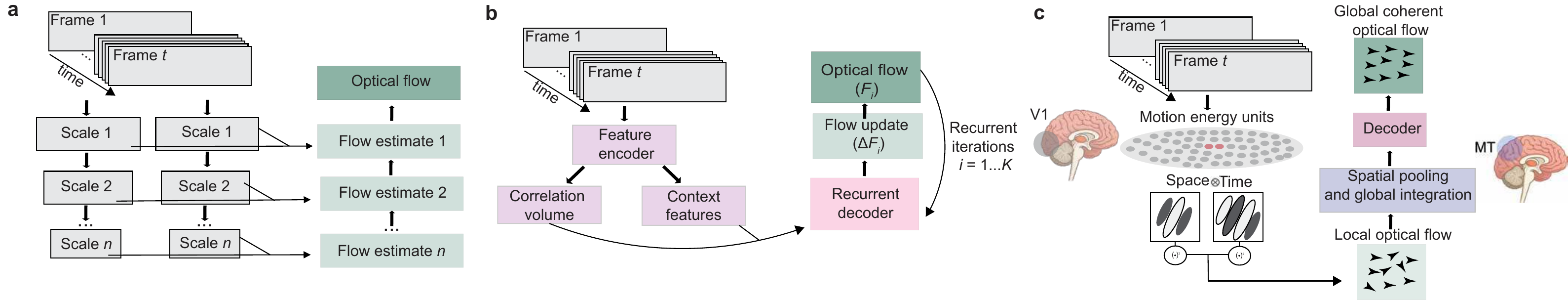}
 \caption{\textbf{Schematic illustrations of three model architecture categories.}
\textbf{(a)} Multi-scale architectures estimate motion using pyramidal feature hierarchies. 
\textbf{(b)} Recurrent-decoding architectures iteratively refine optical flow over multiple update steps. 
\textbf{(c)} Bio-inspired architectures incorporate motion-energy units, or integration mechanisms motivated by primate visual motion pathways.}
  \label{fig:supp_model_schematic}
\end{figure*}

We evaluated ten representative motion-estimation models (Fig.~\ref{fig:fig2}; see Appendix 1 for model details). To ensure a fair comparison, all models were evaluated using the official public implementations and pretrained weights released by their respective authors \cite{pwcnet,lfn2, raft, ccmr, flowdiffuser, ffv1mt, dorsalnet, meattention, sun2025machine}. Collectively, these models span three major architectural categories: multi-scale architectures, recurrent-decoding architectures, and bio-inspired architectures (Fig.~\ref{fig:supp_model_schematic}, Table~\ref{tab:supp_motion-models}). 

\vspace{-11px}
\paragraph{Inference protocol.} Except for DorsalNet, all models were evaluated using their native inference settings: input frames were resized to each model’s required resolution, and the number of frames matched the configuration used
during training. No fine-tuning was performed. Each model outputs dense optical flow fields, from which we derive motion–direction estimates. DorsalNet differs from the other models in that it was originally trained to predict self-motion parameters rather than optical flow. To obtain flow estimates, a linear decoder was trained to map multiscale DorsalNet features from multiple layers into dense optical flow via pointwise temporal projections.

\begin{figure*}[b!]
  \centering
  \includegraphics[width=\linewidth,height=0.72\textheight,keepaspectratio]{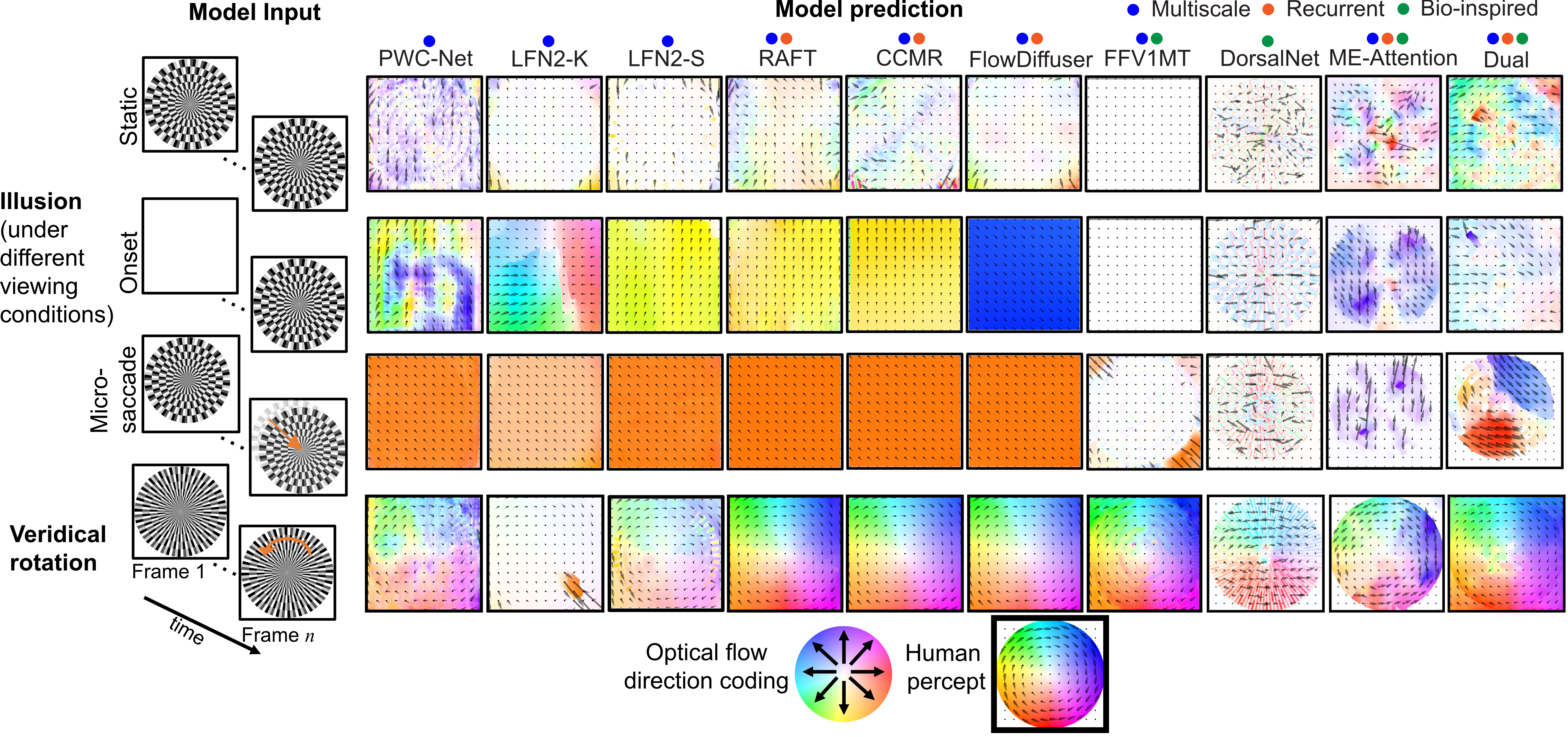}
  \caption{\textbf{Visualization of normalized model-predicted optical flow for grayscale stimuli across central simulated viewing conditions.} Static (top row), onset (second row), and microsaccadic 30-pixel shift (third row). Colored dots correspond to model architectures listed in Appendix 1. The fourth row shows model responses to rotated control stimuli, allowing qualitative comparison with veridical motion. Under static and stimulus-onset conditions, optical flow maps differed qualitatively from the microsaccade condition; however, all models failed to reproduce continuous motion consistent with human perception, with only \emph{Dual} partially capturing counterclockwise motion. Most models accurately captured veridical motion, except \emph{LFN2-K}, which predicted minimal optical flow apart from a small region of strong rightward motion. Among non–bio-inspired models, those with recurrent architectures more effectively captured veridical motion qualitatively.
} 
    \vspace{-7mm}
\label{fig:fig2}
\end{figure*}
\subsection{Viewing Conditions Simulation}

To evaluate how motion-estimation models respond to illusion stimuli under conditions analogous to human viewing, we generated short frame sequences simulating four viewing conditions: \emph{static},
\emph{stimulus onset}, \emph{microsaccade}, and \emph{saccade} (Fig.~\ref{fig:fig2}). All sequences were rendered at $1506 \times 1506$ pixels, with each illusory or control disk centered on the canvas and surrounded by a 120-pixel uniform margin.

\vspace{-11px}
\paragraph{Static.}
The static condition consisted of identical frames across time. Each sequence was constructed by repeatedly presenting the same illusory or
control image for the entire duration of the input window required by the model.

\vspace{-11px}
\paragraph{Stimulus onset.}
To approximate transient responses evoked by stimulus appearance, the first several frames were set to a blank white field, followed by repeated presentation of the same illusory or control image. Only a single change in frame content occurs, analogous to the moment of stimulus onset.

\vspace{-11px}
\paragraph{Microsaccade and saccade.}

To simulate the retinal-image shifts caused by eye movements, we translated the stimulus along a straight trajectory. Two possible directions were used: bottom-right to top-left, or top-left to bottom-right. In each sequence, the image was shifted by a fixed displacement $\Delta$ applied equally in the $x$ and $y$ directions. We tested $\Delta \in \{15,\, 30,\, 60,\, 90,\, 120\}$ pixels, spanning typical human eye movements: microsaccades ($<1^\circ$ visual angle) and saccades (1--8$^\circ$ visual angle)~\cite{otero2008saccades}. Under a standard psychophysics setup with 50~px/$1^\circ$ visual angle, as in prior experiments where human observers reported perceived motion to evaluate model alignment~\cite{sun2025machine}, these shifts cover the range of human eye movements and enable model--human comparisons of motion perception.  Consistent with human oculomotor statistics, we allowed at most three changes in disk location within a sequence, corresponding to the 1--3 microsaccades or saccades typically observed per second~\cite{dimigen2009human} (all model training movies are shorter than one second). See Appendix~4 for analyses of the effects of shift timing and direction on the results, and Appendix~8 for movie demonstrations.
\vspace{-11px}
\paragraph{Peripheral viewing.}
Simulating realistic peripheral viewing requires high-resolution retinal input, whereas the models under study operate on substantially lower
resolutions. As a result, peripheral-viewing effects cannot be accurately reproduced within the native input constraints of these models. For completeness, we provide an approximate implementation in Appendix 3, but peripheral-viewing results are not analyzed in the main text.

\subsubsection{Quantitative Evaluation of Alignment with Expected Human Motion Perception}

To compare results from different models, we generated the motion flow field which mimics the human motion perception by rotating each pixel around the stimulus center (Fig.~\ref{fig:fig2}).This target reflects human perception of the Rotating Snakes illusion, as psychophysics studies show that observers experience coherent, rigid rotational motion in these stimuli~\cite{atala-gerard_rotating_2017, hisakata2018spatial}. The flow magnitude decayed smoothly with radial distance according to $(r/R)^{\gamma} M$, where $r$ is the distance to the center, $R$ is the disk radius, $M$ is the flow magnitude at the boundary, and $\gamma=1$ controls radial decay. This produces a continuous, physically plausible rotational flow field consistent with rigid rotation while avoiding abrupt magnitude discontinuities. We used it to mimic human perception across all viewing conditions, including saccades. The human percept optical flow does not include translational displacement because illusory rotational component is the motion of interest, and the illusory rotation is perceived \emph{after} the eye movement is completed rather than during it~\cite{Otero-Millan6043}. 

For each model, we compare its predicted flow $\mathbf{P}\in\mathbb{R}^{n\times 2}$ to the veridical rotation flow $\mathbf{R}\in\mathbb{R}^{n\times 2}$ at $n$ spatial locations. We quantify alignment using a normalized correlation:
\begin{equation}
\rho(\mathbf{P},\mathbf{R}) 
= 
\frac{
\sum_{i=1}^{n} 
\langle \mathbf{P}_{i,:},\, \mathbf{R}_{i,:} \rangle
}
{\|\mathbf{P}\|_{F}\,\|\mathbf{R}\|_{F}},
\label{eq:flow-corr}
\end{equation}
where $\langle \cdot,\cdot \rangle$ is the standard inner product on $\mathbb{R}^2$ and 
$\|\cdot\|_{F}$ is the Frobenius norm. The resulting score lies in $[-1,1]$:  
\begin{itemize}[noitemsep, topsep=2pt]
    \item $\rho=1$ indicates perfect directional alignment;  
    \item $\rho=0$ indicates orthogonal or uncorrelated flow fields;  
    \item $\rho=-1$ indicates flow in the opposite (clockwise) direction.  
\end{itemize}

We adopt this correlation metric because it emphasizes directional consistency over exact magnitude matching, which is critical for analyzing anomalous motion illusions. For completeness, we also report standard optical-flow metrics including average angular error and endpoint error in Appendix 5.

\section{Results and Analysis}
\subsection{Rotating Snakes}
\subsubsection{Model perception under simulations}

\begin{figure*}[b!]
  \centering
  \includegraphics[width=\linewidth,height=0.72\textheight,keepaspectratio]{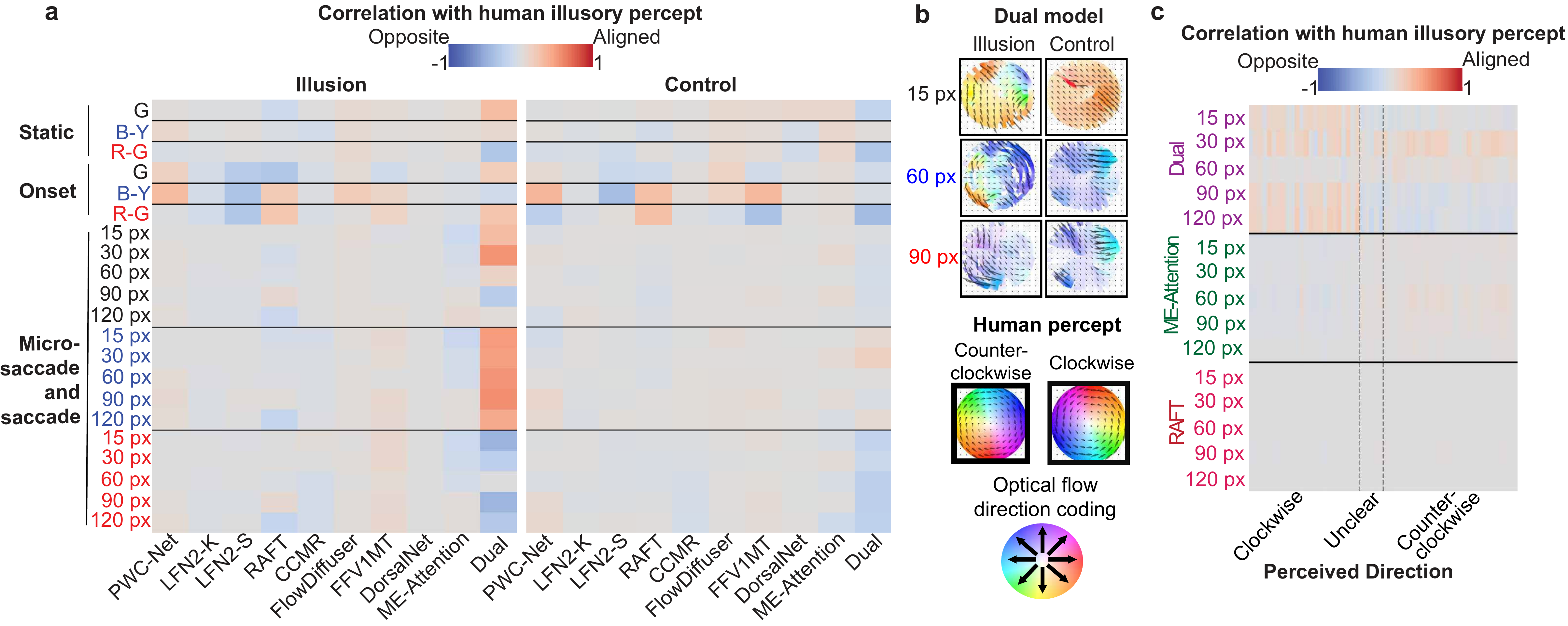}
  \caption{\textbf{Correlation between model-predicted and expected human illusory percepts.} Color types: grayscale (G), blue–yellow (B–Y), and red–green (R–G).
\textbf{(a)} Correlations are shown for illusion (left) and control (right) stimuli under viewing conditions: static, onset, microsaccade, and saccade (all shift magnitudes included). Among all models, the Dual model exhibits the strongest alignment with human illusory percepts for the grayscale and blue–yellow variants, but shows negative correlations for the red–green variant. For control stimuli, Dual produces little to no consistent alignment. 
\textbf{(b)} Optical flow predictions from Dual for illusion and control stimuli at three example combinations of color type and shift magnitude: (G, 15~px), (B–Y, 60~px), and (R–G, 90~px).
\textbf{(c)} Correlation with human behavioral data from~\cite{atala-gerard_rotating_2017}, collected using an independent stimulus set in which the intermediate luminance levels of dark gray and light gray are parametrically varied to modulate the perceived direction and strength of rotation. Stimuli are sorted by perceived direction (clockwise, unclear, counterclockwise). }
  \label{fig:fig3}
\end{figure*}

Figures~\ref{fig:fig2} and~\ref{fig:fig3} summarize
how motion-estimation models respond to the anomalous-motion illusion under viewing simulations (see Appendix 2 for additional results).

\vspace{-11px}
\paragraph{Veridical rotation responses.} We first evaluated whether models can produce correct rotational flow for veridical counterclockwise rotation (Fig.~\ref{fig:fig2}, bottom
row). Most models generated coherent rotational flow fields with the
correct direction, confirming that their architectures can in principle represent circular motion. The only exception was \emph{LFN2-K}, whose flow fields remained nearly zero apart from a small region of rightward motion.

\vspace{-11px}
\paragraph{Qualitative behavior across viewing conditions.} For illusion stimuli, model behavior differed substantially across
viewing conditions (Fig.~\ref{fig:fig2}). Under static input, non–bio-inspired models (PWC-Net, LFN2-S, RAFT, CCMR, FlowDiffuser) produced little or no meaningful flow. Under onset conditions, several models exhibited structured but inconsistent transients unrelated to human percepts. During saccadic shifts, predictions were dominated by the direction of the imposed displacement. 

Among bio-inspired models, FFV1MT produced almost no motion in all conditions, while DorsalNet produced noisy, spatially irregular flows. Although previously shown to capture reverse-phi effects, ME-Attention generated primarily vertical flow. None of these models produced smooth, continuous counterclockwise rotation resembling human perception. Only \emph{Dual} showed partially coherent patterns, especially under microsaccade conditions.

\vspace{-11px}
\paragraph{Quantitative model--human alignment.} The correlation heatmap reveals the same architecture- and condition-dependent trends (Fig.~\ref{fig:fig3}a, left panel). Correlations remained near zero for most models across color variants (grayscale, blue--yellow, red--green), indicating little directional alignment with human-perceived rotation. 

\vspace{-11px}
\paragraph{Dual model: qualitative and quantitative convergence.}
Dual achieved the highest positive correlations for grayscale and blue--yellow illusions under microsaccade simulations (Fig.~\ref{fig:fig3}a, left panel), whereas red--green stimuli often produced negative correlations, indicating rotation opposite to human perception. Importantly, control stimuli in grayscale and blue--yellow (Fig.~\ref{fig:fig3}a, right panel) produced correlations near zero under the the same simulated conditions, confirming that the positive correlations in the illusion case are not explained by the shift trajectory alone, but instead depend on the illusion’s luminance asymmetries. Results for stimuli inducing clockwise illusory motion were symmetric with respect to rotational direction, confirming that model responses reflect the luminance structure of the stimulus rather than a directional bias (Fig.~\ref{fig:supp_fig_clockwise}). Qualitative examples (Fig.~\ref{fig:fig3}b) show that, though not completely continuous, Dual generates globally counterclockwise flow that closely resembles the veridical rotational field for the blue-yellow illusion stimuli at a shift magnitude of 60~px. On control stimuli, Dual’s predictions lack correctly oriented tangential flow, matching near-zero correlations.
 We further evaluated the models against human behavioral data from~\cite{atala-gerard_rotating_2017}, collected using a different set of stimuli. We replicated their stimulus set, in which the intermediate luminance levels $g_1$ (dark gray) and $g_2$ (light gray) are parametrically varied to modulate the perceived direction and strength of the illusion. We also generated optical-flow fields corresponding to human percepts based on participant reports from their experiments. We then compared the model-predicted flow for this independent stimulus set with these human-percept optical-flow fields. Consistent with Fig.~\ref{fig:fig3}a, Dual shows the strongest alignment with human perception across stimulus variants (Fig.~\ref{fig:fig3}c), especially under the 30~px microsaccade condition, although the correlations are not uniformly positive. This suggests that Dual's alignment generalizes beyond the stimuli used in our main experiments.

\begin{figure*}[ht]
  \centering
  \includegraphics[width=\linewidth,height=0.72\textheight,keepaspectratio]{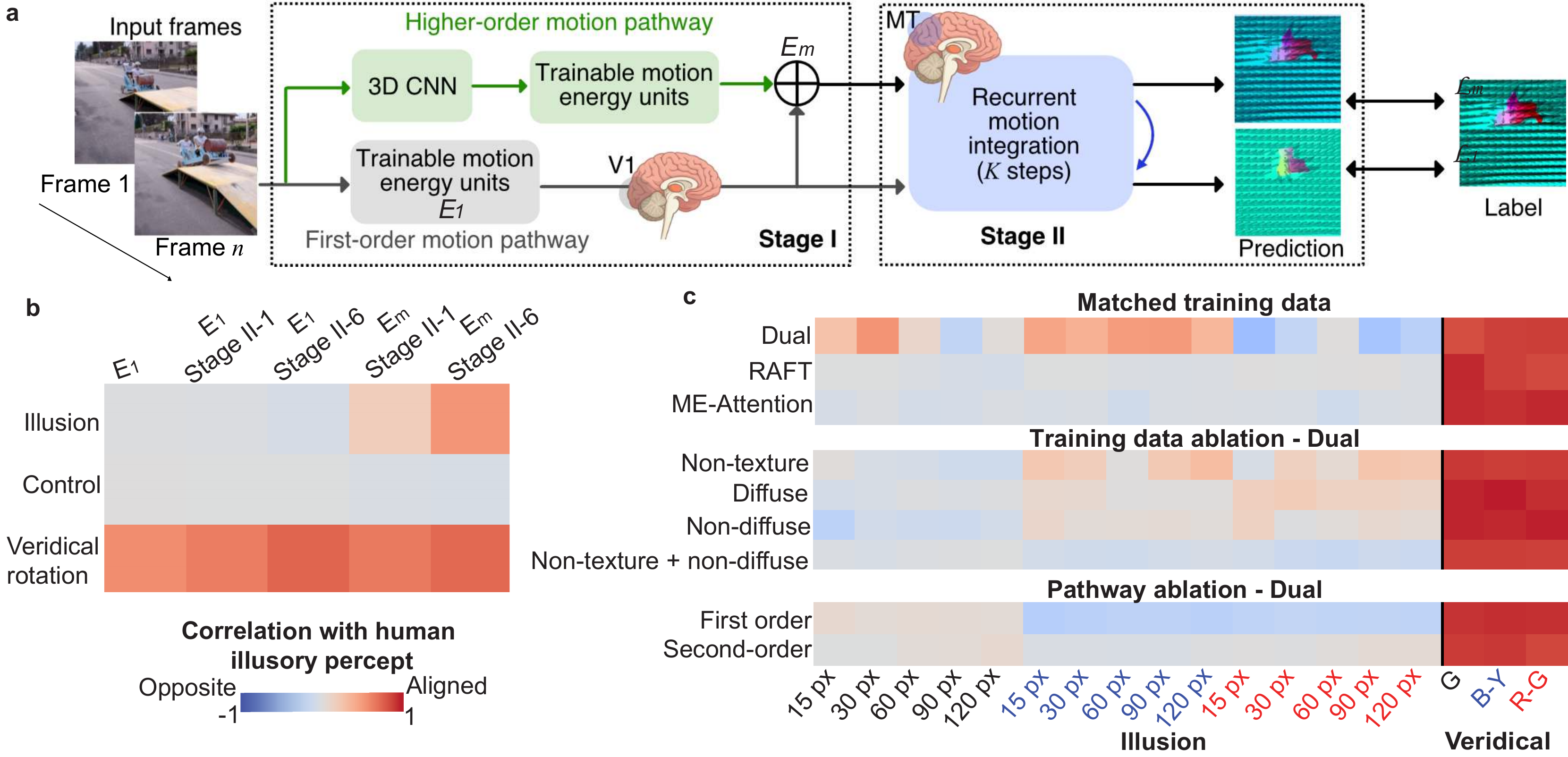}
  \caption{\textbf{Ablation analysis of the Dual model to identify the critical components for reproducing human illusory motion percepts.}
\textbf{(a)} Schematic of the Dual architecture showing the first-order ($E_{1}$) and higher-order ($E_{2}$) motion pathways, their fusion into $E_{m}$, and subsequent recurrent integration (Stage~II).
\textbf{(b)} \textbf{Probing analysis.} Optical flow predictions for the grayscale illusion, control, and veridical rotation stimuli under a 30-pixel microsaccade. Columns show outputs from: the first-order pathway alone ($E_{1}$), the first and last recurrent
iterations applied to $E_{1}$ (Stage~II-1 and Stage~II-6), and the corresponding first and last recurrent iterations applied to $E_m$. 
\textbf{(c)} \textbf{Control and ablation analysis.} Correlations are shown for Dual, RAFT, and ME-Attention trained on the full dataset, as well as for Dual trained with dataset and architectural ablations, evaluated on illusion and veridical-rotation stimuli.
}
  \label{fig:fig4}
\end{figure*}

\subsubsection{Probing, Control and Ablation Analyses for Dual Model}

The Dual architecture (Fig.~\ref{fig:fig4}a) processes motion through two parallel pathways prior to recurrent integration. The \emph{first-order channel} ($E_1$, analogous to V1 in primate visual cortex) applies spatiotemporally separable Gabor filters across 8 spatial scales to grayscale input, producing 256-dimensional motion-energy maps at $96\times96$ resolution. The \emph{higher-order channel} ($E_2$) extracts non-Fourier motion features from RGB sequences using five layers of 3D CNNs, followed by trainable Gabor filtering to apply motion energy constraints, also outputting 256-dimensional feature maps. Both channels undergo the same normalization process, after which they are merged via concatenation and a $1\times1$ convolution, with the resulting fused output $E_m \in \mathbb{R}^{H/8 \times W/8 \times 256}$ squared and fed to stage II. Stage II consists of a 6-layer transformer module (analogous to MT in primate visual cortex) with graph-based attention. A linear decoder produces optical flow at every recurrent iteration.

During training, the $E_1$ pathway is decoded both before recurrence (Stage~I) and after each recurrent iteration (Stage~II-$k$). The fused representation $E_m$ is not decoded at Stage~I, but its features after each recurrent iteration (Stage~II-$k$) are decoded
and supervised. Let $\mathrm{flow}_{E_1}^{\mathrm{bi}}$
denote the sequence of predicted flow fields obtained from $E_1$, consisting of the direct Stage~I prediction ($k=0$) and the $K$ recurrent Stage~II predictions ($k=1,\dots,K$). Likewise, let $\mathrm{flow}_{E_m}^{\mathrm{bi}}$ denote the sequence of flow fields decoded from the fused pathway $E_m$ after each of the $K$ recurrent iterations. Each predicted flow field is compared to the ground-truth optical flow $\mathbf{u}^{*}$. The total training loss is the sum of the sequence losses:
\begin{equation}
\mathcal{L}
  = \mathrm{loss}\!\left(\mathrm{flow}_{E_1}^{\mathrm{bi}},\,
                                   \mathbf{u}^{*}\right)
  + \mathrm{loss}\!\left(\mathrm{flow}_{E_m}^{\mathrm{bi}},\,
                                   \mathbf{u}^{*}\right).    
\end{equation}

\vspace{-11px}
\paragraph{Probing analysis.}
 To localize where illusory-motion signals emerge during processing, we decoded optical flow from intermediate representations at each pathway and recurrent stage without modifying the model weights (Fig.~\ref{fig:fig4}b). For veridical-rotation stimuli, flow fields decoded from all stages show high correlations with the counterclockwise rotational target. In contrast, for illusion stimuli, flow fields decoded from $E_1$ alone remain near zero even with recurrent processing (Fig.~\ref{fig:fig4}b), indicating that the first-order channel alone is insufficient to generate rotation-consistent signals. Introducing the higher-order channel substantially changes model behavior: the fused representation $E_m$ at Stage~I yields higher correlations for illusion stimuli. Recurrent integration further amplifies this alignment. Correlation values peak at Stage~II-6 for both illusion and veridical-rotation stimuli, while control stimuli remain near zero throughout, confirming that this alignment is driven by the luminance structure of the illusion stimuli.

\vspace{-11px}
\paragraph{Control Analysis.}
To identify the contributions of model architecture to illusory-motion alignment, we retrained RAFT and ME-Attention on the same training data as Dual. We selected ME-Attention as a comparison because of its architectural similarity to Dual, differing primarily in the absence of the second-order channel, and RAFT as an additional comparison because of its recurrent decoding. After matching training data, both models still show substantially weaker correspondence with human illusory percepts than Dual while still producing accurate optical flow for veridical rotation, suggesting that architectures play a key role beyond training data.(Fig.~\ref{fig:fig4}c - top panel).

\vspace{-11px}
\paragraph{Ablation analysis.}
We retrained Dual with subsets of the training data and ablation of first- and second-order pathways; all variants produced accurate optical flow for veridical rotation, confirming that neither manipulation impairs basic motion estimation. Removing each training-data subset individually reduces correspondence, whereas jointly removing the non-textured and non-diffuse object motion subsets largely eliminates it (Fig.~\ref{fig:fig4}c, middle panel). Removing either the first-order or higher-order pathway likewise substantially reduces model--human correspondence, indicating that both training-data composition and dual-pathway architecture are necessary for improved model--human correspondence (Fig.~\ref{fig:fig4}c, bottom panel).

See Appendix 6 for qualitative examples of model-predicted flow across probing, control, and ablation analyses. 

\subsubsection{Unit-level Analysis for Dual Model}
To visualize how illusion stimuli activate local motion-energy units, we analyzed $E_1$, which consists of spatiotemporally separable Gabor filters sensitive to local luminance-defined motion. Each unit's preferred direction and spatiotemporal frequency were determined by testing drifting Gabor stimuli at twelve uniformly sampled directions and $8\times8$ logarithmically sampled spatiotemporal frequency combinations; the direction-frequency pair yielding the largest response standard deviation was selected as that unit's preferred direction and spatiotemporal frequency~\cite{sun2025machine}.  To identify units most sensitive to rotational motion, we ranked units by their absolute activation difference between veridical rotation and static control, selecting the top 25\% and retaining those more strongly activated by rotation than by the static control.

\begin{figure}[t]
  \centering
  \IfFileExists{final_fig7.png}{%
    \includegraphics[width=0.5\linewidth,height=0.5\textheight,keepaspectratio]{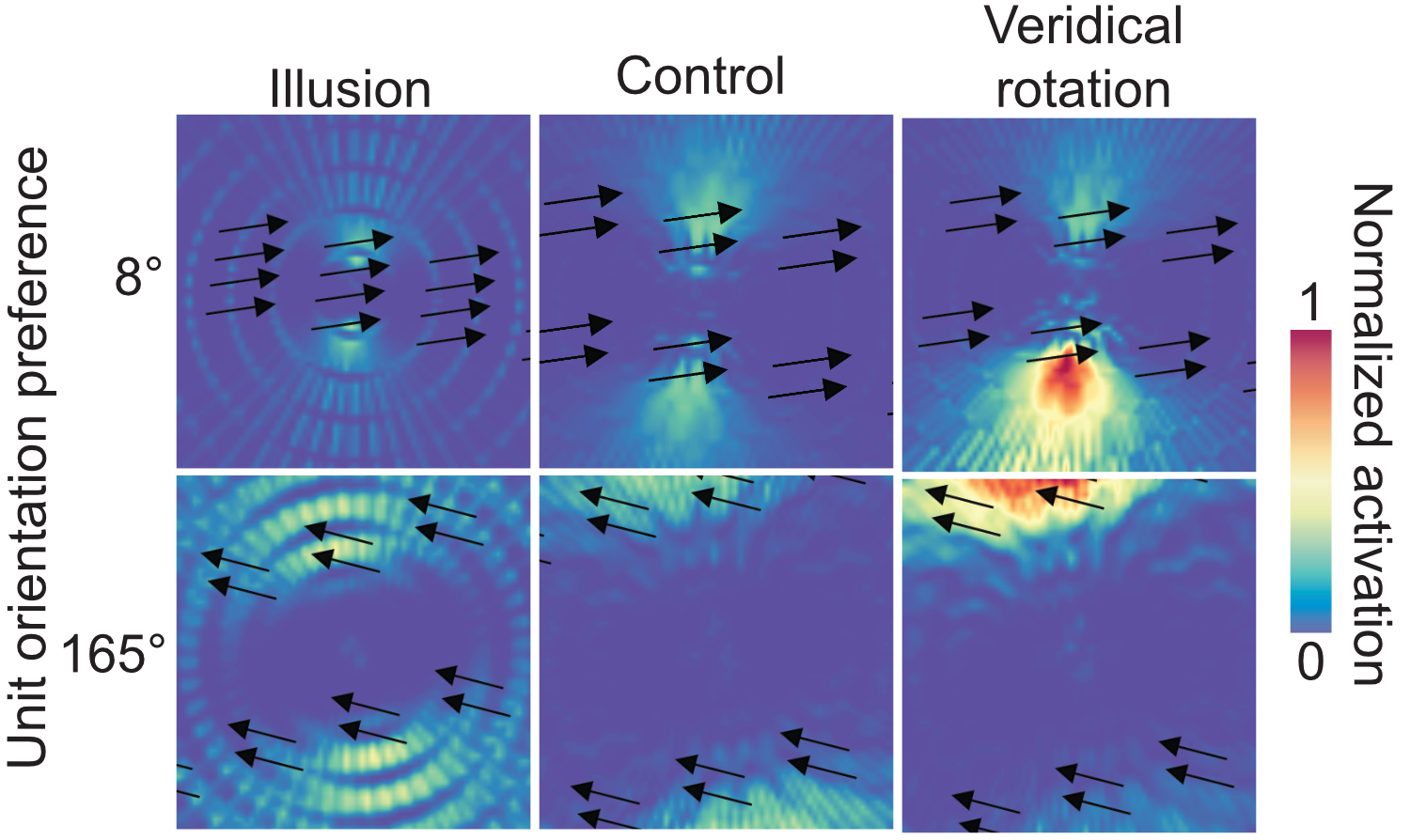}%
  }
  \par\medskip
  \caption{\textbf{Spatial response profiles of two exemplar $E_{1}$ units sensitive to rotational motion.} Activations are normalized per unit to highlight spatial structure. For each unit (rows), spatial activation maps are shown for the illusion, control, and veridical rotation conditions. The preferred motion direction
($\theta$) of each unit is indicated by the black arrow overlaid on each map.}
  \label{fig:fig5}
\end{figure}

\vspace{-11px}
\paragraph{Observations.}
Figure~\ref{fig:fig5} visualizes spatial activation patterns of two representative $E_1$ units in response to the illusion, control, and veridical rotation conditions. For both units, although the illusion condition produces a detectable activation peak that roughly aligns with the rotation direction, this pattern is substantially more diffuse than in either the control or veridical rotation conditions.  Veridical rotation evokes a pronounced asymmetric activation profile consistent with true rotational flow. In contrast, illusion and control stimuli elicit  more diffuse, spatially symmetric patterns  insufficient to form a continuous rotation.  This suggests that higher-order features or recurrent integration are required to generate the global rotation observed in the full Dual model.

\subsection{Related Anomalous Motion Illusions}
We further evaluated each model on peripheral drift illusions (PDIs), central drift illusions (CDIs), and the Ouchi illusion (Fig.~\ref{fig:fig6}; see Appendix~7 for details). Human observers perceive clockwise rotation in PDIs and CDIs, and relative center--surround sliding in the Ouchi pattern.

For peripheral drift illusions (Fig.~\ref{fig:fig6}a), ME-Attention and Dual generated weak, noisy, or oppositely oriented motion signals; only the blue--yellow variant elicited partial alignment in Dual. RAFT produced almost no flow. For central drift illusion (Fig.~\ref{fig:fig6}b), no model captured the expected slow rotational 'creeping' motion percept: RAFT produced minimal flow while ME-Attention and Dual generated locally structured but globally incoherent patterns.

Humans perceive the central region in the Ouchi illusion image as sliding relative to the surround when eye movements occur. ME-Attention and Dual produced pronounced segmentation-like responses that separate center and surround with different motion speed between these two regions (Fig.~\ref{fig:fig6}c). In contrast, RAFT produced
low-amplitude, spatially structured flow that did not correspond to the perceived figure–ground segregation.
\begin{figure}[t]
  \centering
  \includegraphics[width=0.5\linewidth,height=0.5\textheight,keepaspectratio]{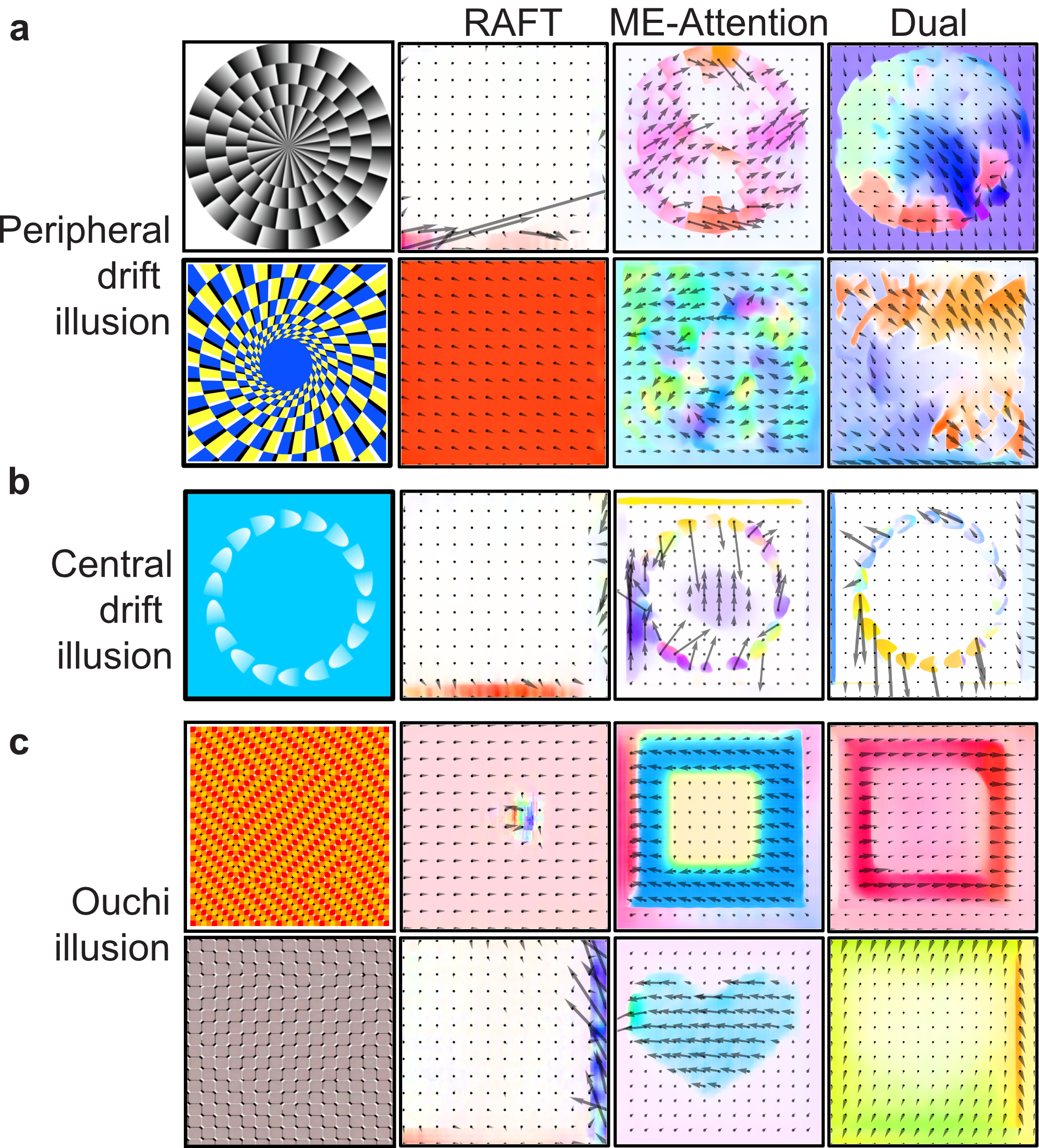}
\par\medskip
\caption{\textbf{Optical flow predictions for other anomalous-motion illusion stimuli.} (\textbf{a}) Peripheral drift illusion: observers typically experience smooth clockwise rotation following a blink or small eye movement. (\textbf{b}) Central drift illusion: the white elements appear to rotate slowly in a clockwise direction. (\textbf{c}) Ouchi illusion: eye movements induce a relative sliding motion
between the center and surround. Representative outputs from RAFT, ME-Attention, and Dual are shown under the microsaccade simulation.}
  \label{fig:fig6}
\end{figure}

\section{Discussion}
In this study, we investigated whether contemporary optical-flow models reproduce the illusory rotation perceived by humans when viewing static images. Across a diverse model suite spanning classical, diffusion- or transformer-based, and biologically inspired architectures, no model
fully recovered the continuous rotational motion reported by human observers. Human-like responses emerged only under certain simulated viewing conditions (primarily microsaccades), and even then were fragmentary and spatially irregular. Among all models tested, Dual exhibited the closest correspondence to human perception, but remained far from replicating the smooth, coherent rotation seen in human behavior.

\vspace{-11px}
\paragraph{Implications for Model Design and Training Data.}
Our findings highlight the importance of biologically inspired components for capturing illusory motion. More broadly, they suggest two promising directions for improving machine motion perception. First, current models rarely incorporate the eye movements that dominate human retinal input. Recent work shows that simulating microsaccades or
naturalistic retinal slip can preserve fine detail and stabilize motion estimation~\cite{He2024AMI, Wu2024RGCJitter}. Incorporating realistic fixational statistics into training may therefore yield more robust optical-flow models.

Second, the success of neuroscience-inspired models in other challenging vision tasks supports the value of integrating human-like motion processing. For example, recent work on figure–ground segmentation shows that deep networks equipped with motion-energy front ends outperform state-of-the-art optical-flow methods in challenging stimuli such as random-dot segmentations~\cite{tangemann2024object}. We also demonstrate that models with first-order motion-energy mechanisms successfully segment Ouchi illusory images that are often believed to be “human-only” perceptual discriminations. These findings point toward more robust machine vision by combining bio-inspired architectures.
\vspace{-11px}
\paragraph{Neural mechanisms of anomalous motion illusions.} From a scientific perspective, our analyses provide insight into the computations that may underlie different classes of anomalous-motion illusions. 

For the Rotating Snakes stimulus, ablations of the Dual model reveal that it is best reproduced by an architecture combining first-order, higher-order, and recurrent mechanisms. This stands in partial tension with existing computational models, which are predominantly first-order, attributing the illusion to contrast asymmetries that produce latency differences activating motion detectors~\cite{conway_neural_2005, backus_illusory_2005, bach_rotating_2020, faubert1999peripheral, fermuller_illusory_2010}. That purely first-order models (\eg, FFV1MT, ME-Attention) fail suggests that first-order accounts, while capturing an important component, may be incomplete: full correspondence with human perception may additionally require higher-order and recurrent integration, as instantiated in the Dual model.

A previous study showed that training on non-textured and non-diffuse object motion is critical for the emergence of higher-order motion perception~\cite{sun2025machine}. The drastically reduced alignment with human illusory perception after ablating these two data components further suggests the importance of higher-order motion mechanisms. However, how other data components, such as diffuse object motion, affect illusory perception requires further investigation.

For the Ouchi illusion ~\cite{hine1997ouchi}, both Dual and ME-Attention model—each containing an explicit motion-energy front end, produce robust relative-motion signals like humans. This mirrors perceptual evidence that the Ouchi effect can be accounted for by first-order motion-energy processing and biased local motion integration, without requiring higher-order features~\cite{mather2000integration, Spillmann2013}. These results suggest that some anomalous-motion illusions (\eg, Ouchi) may indeed be largely driven by first-order mechanisms, whereas others such as Rotating Snakes place additional demands on higher-order processing.

\vspace{-11px}
\paragraph{Limitations and Future Work.}
First, we analyzed only a subset of anomalous-motion illusions; applying our evaluation framework to broader illusion families~\cite{kitaoka_classify} may reveal additional computational primitives. Second, our simulated viewing conditions approximate but do not fully capture the statistics of human eye movements. Real behavior involves a mixture of microsaccades, ocular drifts, and high-frequency tremor, whose temporal and spatial properties differ substantially from simple image jitter. Incorporating more realistic microsaccade and drift models ~\cite{Engbert2011, Nau2020DriftSim, Allgeier2024FixEMSim} may alter model behavior. Third, existing motion-estimation models tested so far fail to distinguish self-motion from object motion, and our discrete-snapshot simulation omits efference-copy-based spatial remapping~\cite{bays2007remapping}. Fourth, peripheral drift illusions such as Rotating Snakes induce human illusory motion perception when the stimuli appear in the peripheral visual field rather than the fovea ~\cite{faubert1999peripheral}, yet all models failed to replicate this effect, likely because they lack mechanisms for peripheral vision.

In future work, we will develop new bio-inspired motion models that incorporate peripheral-vision mechanisms under realistic eye-movement statistics and temporal windows. Such models may not only improve robustness on challenging vision tasks but also provide mechanistic insight into how the primate visual system transforms retinal image changes into stable motion signals.

\section*{Acknowledgments}
We thank Shin'ya Nishida and Ning Qian, as well as members of the Kriegeskorte Lab, for their helpful insights and comments on our work. 

{
    \small
    \bibliographystyle{ieeenat_fullname}
    \bibliography{main}
}

\clearpage
\appendix
\setcounter{section}{0}
\renewcommand{\thesection}{Appendix~\arabic{section}}
\renewcommand{\thesubsection}{\arabic{section}.\arabic{subsection}}
\renewcommand{\thesubsubsection}{\arabic{section}.\arabic{subsection}.\arabic{subsubsection}}
\setcounter{figure}{0}
\setcounter{table}{0}
\renewcommand{\thefigure}{S\arabic{figure}}
\renewcommand{\thetable}{S\arabic{table}}

\section{Details on model parameters and architectures}
We evaluated ten motion-estimation models (Table~\ref{tab:supp_motion-models}):

\textbf{PWC-Net.} A convolutional neural network (CNN) for optical flow estimation through coarse-to-fine refinement. A cost volume is constructed between the first image’s features and the warped second-image features, which a CNN was trained end-to-end to refine the flow estimate \cite{pwcnet}.

\textbf{LFN2-K and LFN2-S.} \emph{LiteFlowNet2} estimates optical flow by matching local feature neighbors and refining flow through cost-volume correlation and learned regularization, while maintaining a lightweight architecture. The authors provide two trained variants: LFN2-K, fine-tuned for real-world driving scenes (KITTI), and LFN2-S, fine-tuned for rich-textured scenes (Sintel) data \cite{lfn2}.

 \textbf{RAFT.} \emph{Recurrent All-Pairs Field Transforms} extracts per-pixel deep features from two consecutive frames and constructs a multi-scale correlation volume that contains all pairwise feature similarities. A recurrent module then iteratively updates the flow field by looking up values in this correlation volume and performing optimization. The core idea is to maintain the flow at full resolution throughout iterations (instead of coarse-to-fine), enabling it to recover fine details and handle large motions with high fidelity \cite{raft}.

 \textbf{CCMR.} \emph{Context-guided Coarse-to-fine Motion Reasoning} is a recent optical flow model that integrates attention mechanisms into a multi-scale flow architecture. Global context features are first computed to guide localized motion aggregation at each pyramid level. It has demonstrated strong performance in occluded regions \cite{ccmr}.

 \textbf{FlowDiffuser.} A novel approach that reframes optical flow as a conditional generation task using diffusion models. It starts from random noise and progressively denoises it into a flow field, conditioned on the two input frames. \emph{FlowDiffuser} can naturally model uncertainty and avoid some training biases of direct regression \cite{flowdiffuser}.

 \textbf{FFV1MT.} \emph{Feed-Forward V1–MT model} is a human-inspired computational model for optical flow estimation. This feedforward model has two layers that use V1-like spatiotemporal motion-energy filters and MT-like pattern pooling to estimate dense optical flow \cite{ffv1mt}.

 \textbf{DorsalNet.} A goal-driven model of the primate dorsal visual stream. This model is a 3D ResNet that was trained to predict an agent’s self-motion parameters from video input. The core idea is that by learning to estimate how an observer moves through the environment, the network develops internal representations similar to neurons in the brain’s dorsal pathway (which is responsible for motion perception) \cite{dorsalnet}.

 \textbf{ME-Attention.} \emph{Motion Energy–based Attention model} is a two-stage neural model of human motion perception to combine classical motion energy sensing with modern attention mechanisms. The local signals are fed into a recurrent self-attention network that adaptively integrates motion over space and time. Notably, this model reproduces several neurophysiological and psychophysical observations \cite{meattention}.

 \textbf{Dual.} Compared with the ME-Attention model, which uses a single V1--MT motion-energy stream to explain primarily first-order motion perception, the \textit{Dual} model extends this architecture to a dual-pathway system that jointly learns first- and second-order motion. Trained on naturalistic videos with diverse material properties, the Dual model reproduces key psychophysical and neurophysiological findings and achieves dense optical flow and motion segmentation performance comparable to modern computer-vision models \cite{sun2024acquisition,sun2025machine}. 

\begin{table}[ht]
  \caption{Summary of motion estimation models.}
  \label{tab:supp_motion-models}
  \centering
  \begin{tabular}{@{}lccc@{}}
    \toprule
    Model & Parameters & Input Size & \begin{tabular}{@{}c@{}}Input\\Frames\end{tabular} \\
    \midrule
    PWC-Net~\cite{pwcnet} & 8.75 M &  $1024 \times 436$ & 2 \\
    LFN2~\cite{lfn2} & 6.42 M & $1024 \times 436$ & 2 \\
    RAFT~\cite{raft} & 5.26 M & $520 \times 960$ & 2 \\
    CCMR~\cite{ccmr} & 11.5 M & $512 \times 512$ & 2 \\
    FlowDiffuser~\cite{flowdiffuser} & 14.5 M & $512 \times 512$ & 2 \\
    FFV1MT~\cite{ffv1mt} & N.A. & $768 \times 768$ & 5 \\
    DorsalNet~\cite{dorsalnet} & 57.2 M & $768 \times 768$ & 6 \\
    ME-Attention~\cite{meattention} & 14.7 M &  $768 \times 768$ & 11 \\
    Dual~\cite{sun2025machine} & 25.6 M & $768 \times 768$ & 15 \\
    \bottomrule
  \end{tabular}
\end{table}

\section{Additional examples of model predictions}
\begin{figure*}[t]
  \centering
  \includegraphics[width=\linewidth,height=0.72\textheight,keepaspectratio]{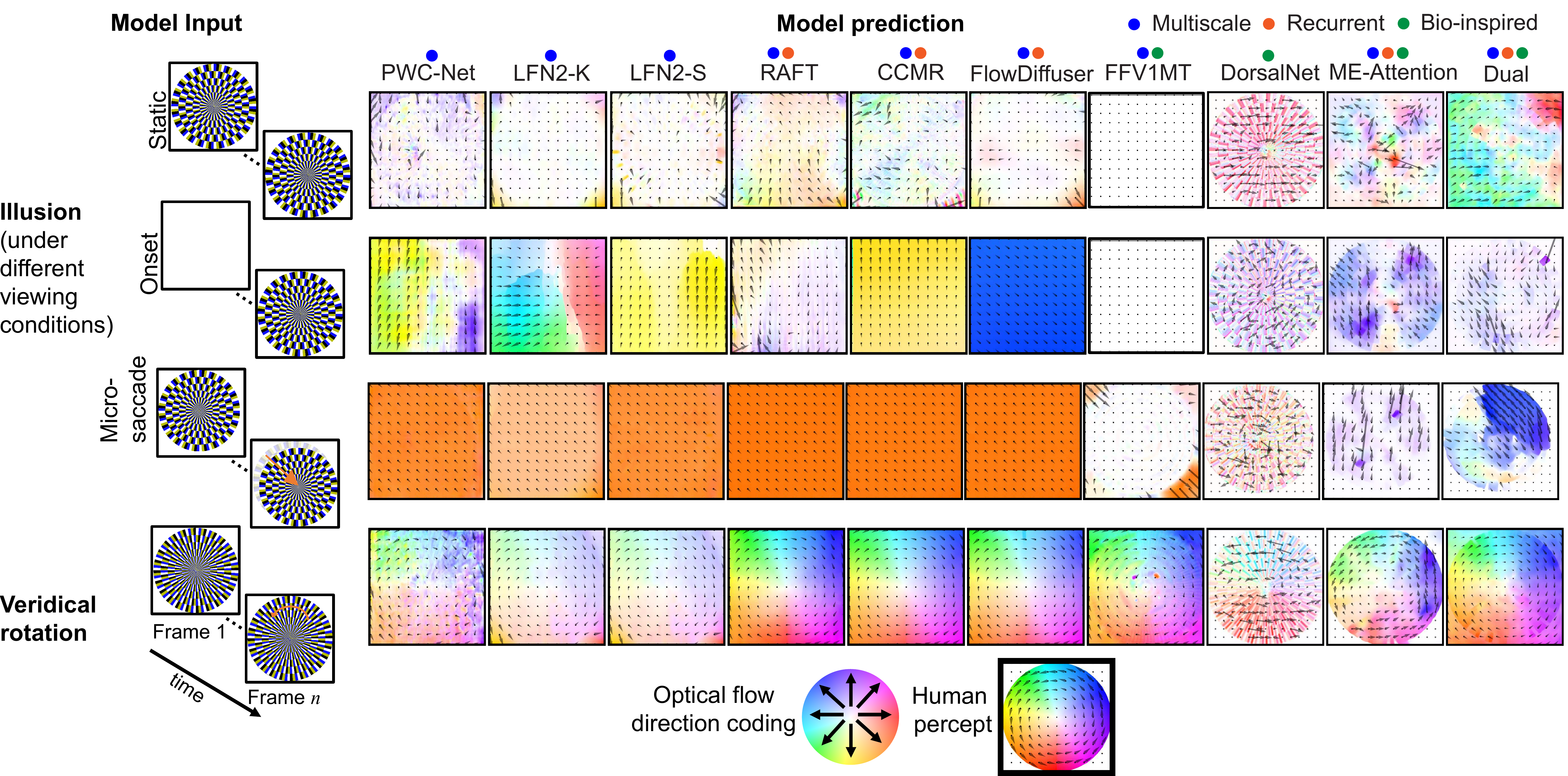}
  \caption{\textbf{Visualization of normalized model-predicted optical flow for blue--yellow illusion stimuli.} We evaluated model predictions under three simulated viewing conditions commonly used in psychophysics: (i) \emph{static} presentation, (ii) \emph{onset} presentation, and (iii) \emph{microsaccade} shifts (small translational displacements that approximate eye movements). For comparison, model predictions for \emph{veridical rotation} of control stimuli are also shown. Colored dots above each column indicate the model architecture type (multiscale, recurrent, or bio-inspired). Optical flow direction is encoded using a circular color wheel (\emph{Optical flow direction coding}, bottom), where hue denotes direction and brightness denotes normalized magnitude. The \emph{human percept} icon (bottom right) represents the expected counterclockwise rotational flow perceived by human observers.} 

  \label{fig:supp_fig1_BY}
\end{figure*}

\begin{figure*}[t]
  \centering
  \includegraphics[width=\linewidth,height=0.72\textheight,keepaspectratio]{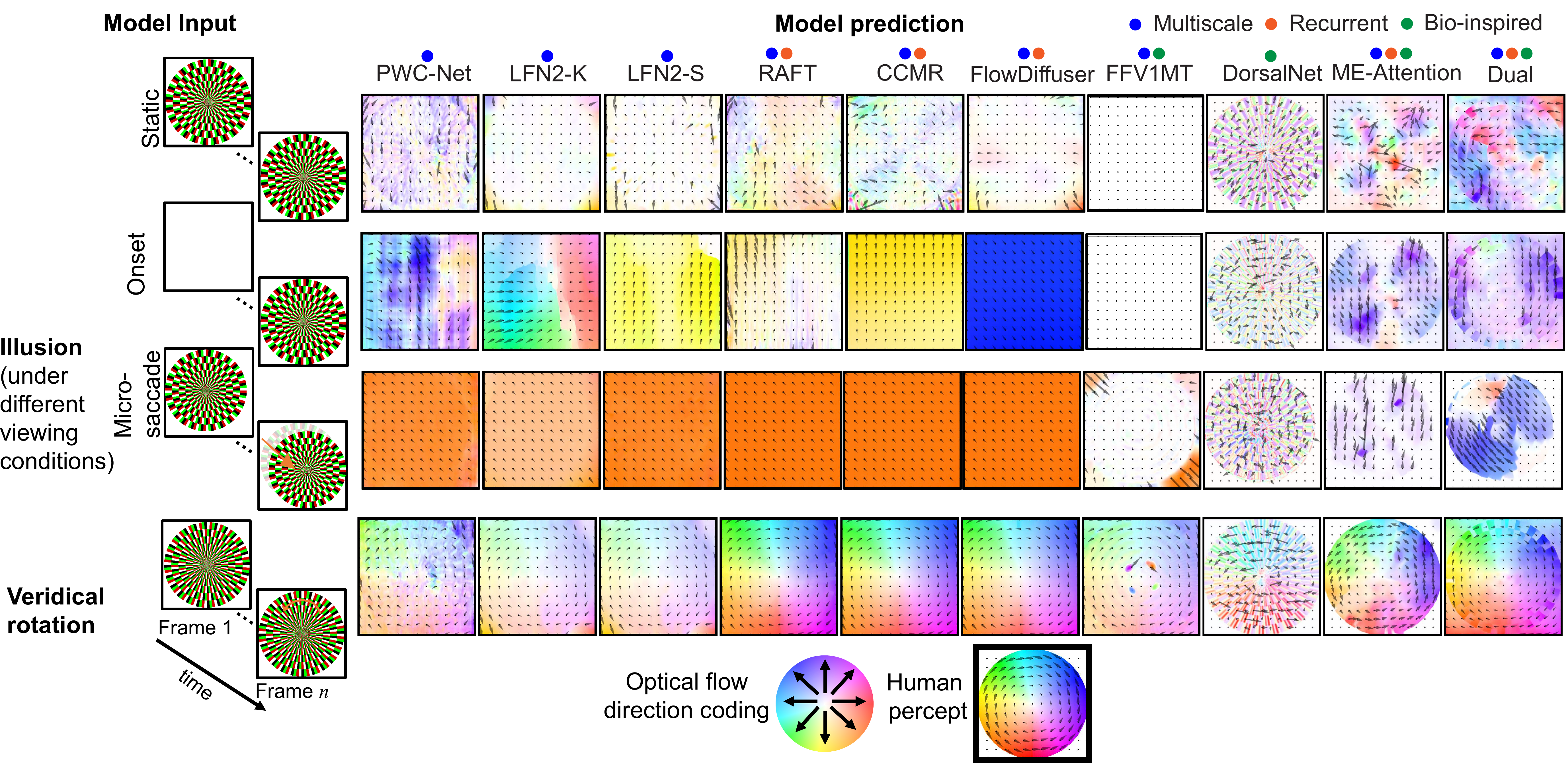}
  \caption{\textbf{Visualization of normalized model-predicted optical flow for red--green illusion stimuli.} Figure layout follow the conventions introduced in Figure~\ref{fig:supp_fig1_BY}. Here we present the corresponding results for the red--green variant under identical simulated conditions.} 
 
  \label{fig:supp_fig1_RG}
\end{figure*}

\paragraph{Blue--yellow and Red--green variants.} The Rotating Snakes illusion induces a robust percept of counterclockwise motion in human observers despite the stimulus being physically static. Figures~\ref{fig:supp_fig1_BY} and ~\ref{fig:supp_fig1_RG} show model-predicted optical flow for the blue--yellow and the red--green variants across viewing conditions. Non–bio-inspired models produce minimal or shift-dominated flow, whereas the Dual model exhibits the clearest rotational components under microsaccades, consistent with the trends shown in the main text. For the red--green variant, Dual often produces flow opposite to the expected direction, mirroring the
negative correlations reported in the main text.

\begin{figure*}[tp]
  \centering
  \includegraphics[width=\linewidth,height=0.68\textheight,keepaspectratio]{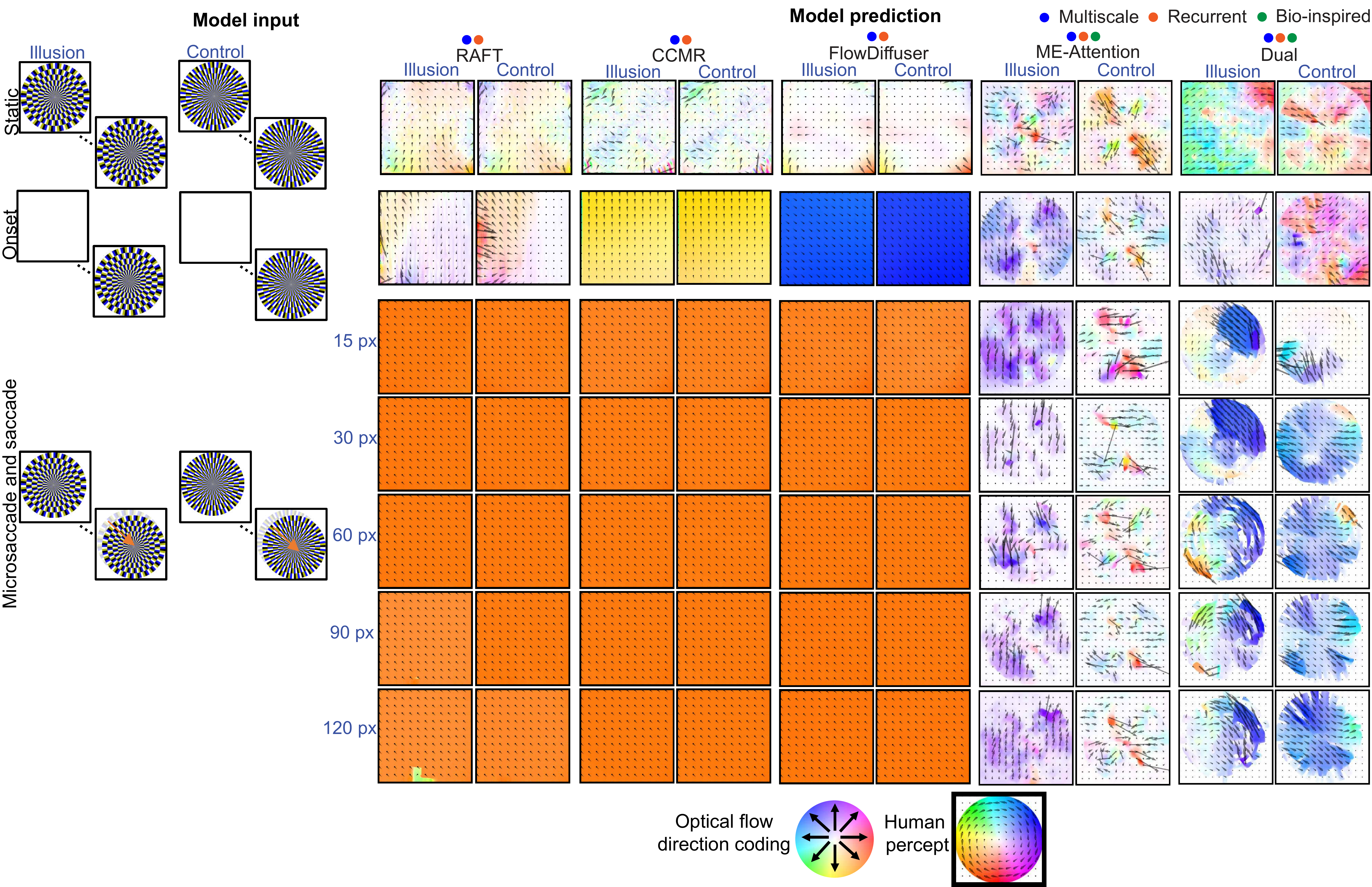}
  \caption{\textbf{Visualization of normalized model-predicted optical flow for blue--yellow illusion and control stimuli across select models.} 
This figure extends the results in Figures~2--3 by directly comparing model outputs for illusion versus control stimuli under matched viewing conditions (static, onset, and microsaccade/saccade shifts of 15--120~px). Results are shown for a subset of representative architectures (RAFT, CCMR, FlowDiffuser, ME-Attention, and Dual). Figure layout, color conventions, and direction-coding icons follow those introduced in Supplementary Figure~\ref{fig:supp_fig1_BY}.}
  \label{fig:supp_fig1_models}
\end{figure*}

\vspace{-11px}
\paragraph{Illusion vs.\ control comparisons.} Figure~\ref{fig:supp_fig1_models} directly compares optical flow estimates for illusion and control stimuli across a representative subset of models and shift magnitudes. For non–bio-inspired architectures (RAFT, CCMR, FlowDiffuser), responses to illusion and control stimuli are nearly identical, indicating that the models do not encode the asymmetric luminance gradients critical to the illusion. ME-Attention shows some sensitivity to local structure, but its flow fields remain noisy and lack global coherence. Only Dual consistently differentiates the illusion from its control, particularly under 60--120~px saccades, generating local flow components that align with the expected counterclockwise rotation. Nonetheless, even Dual fails to reproduce the spatial continuity and direction stability observed in human perception, confirming the main-text conclusion that recurrence and dual-channel processing are necessary but not sufficient for full illusory motion replication.

\begin{figure}[t]
  \centering
  \includegraphics[width=0.6\linewidth,height=0.5\textheight,keepaspectratio]{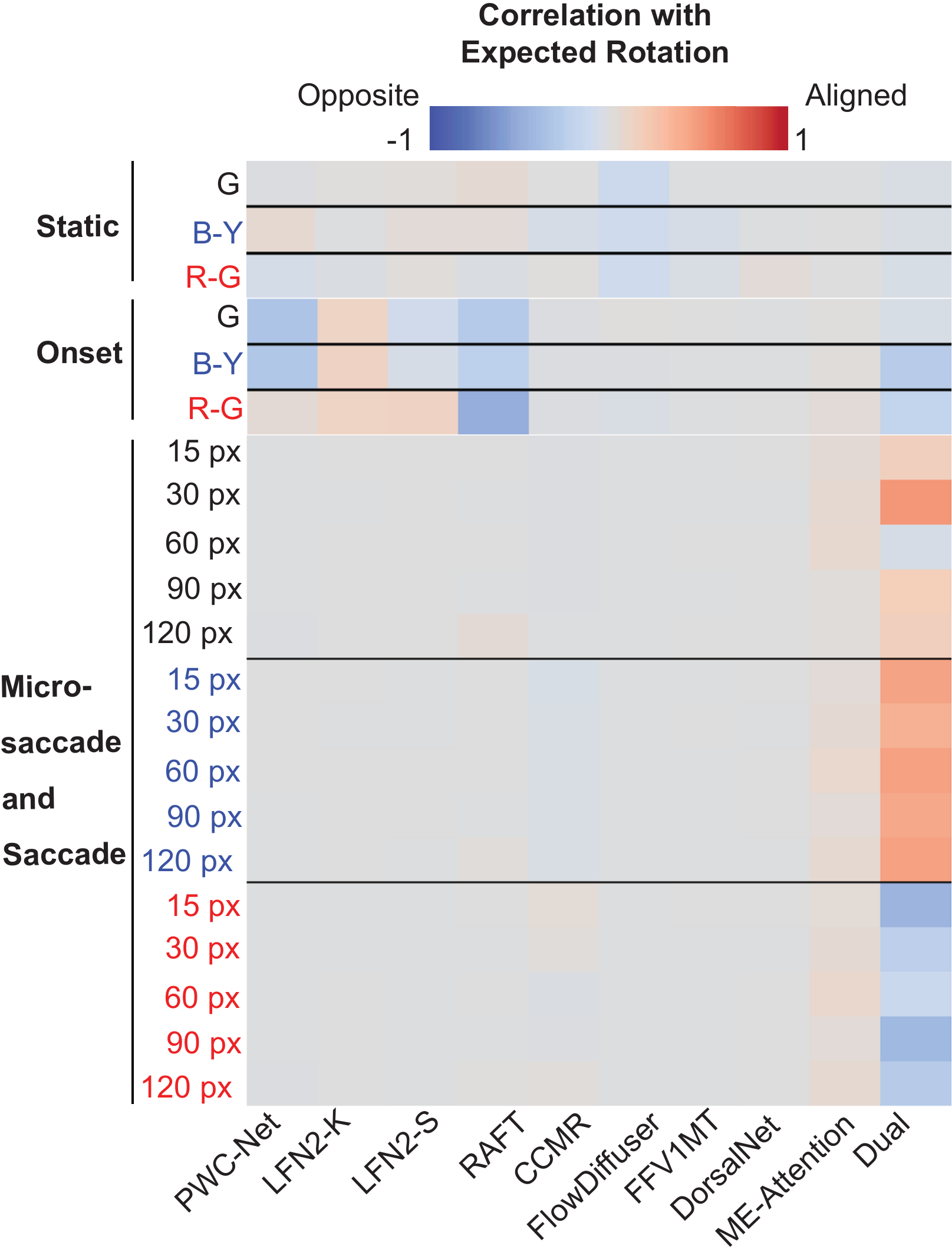}
  \caption{\textbf{Correlation between model-predicted and expected human illusory percepts for illusion stimuli inducing clockwise motion perception, and for corresponding control stimuli across models and viewing conditions.} 
}
  \label{fig:supp_fig_clockwise}
\end{figure}

\vspace{-11px}
\paragraph{Generalization to stimuli inducing opposite illusory motion perception.} To verify that model responses reflect the luminance structure of the stimulus rather than a directional bias, we reversed the luminance order of the micropattern to induce a clockwise percept and re-evaluated all models. Figure~\ref{fig:supp_fig_clockwise} shows model-predicted optical flow for stimuli that induce clockwise illusory motion and their corresponding controls. Correlations between model predictions and the expected human illusory percept remain consistent across clockwise and counterclockwise conditions 
(Figure~\ref{fig:supp_fig_clockwise}), confirming that the results reported in the main text are not specific to a particular rotation direction.

\newpage
\section{Peripheral viewing}
\begin{figure*}[tp]
  \centering
  \includegraphics[width=\linewidth,height=0.72\textheight,keepaspectratio]{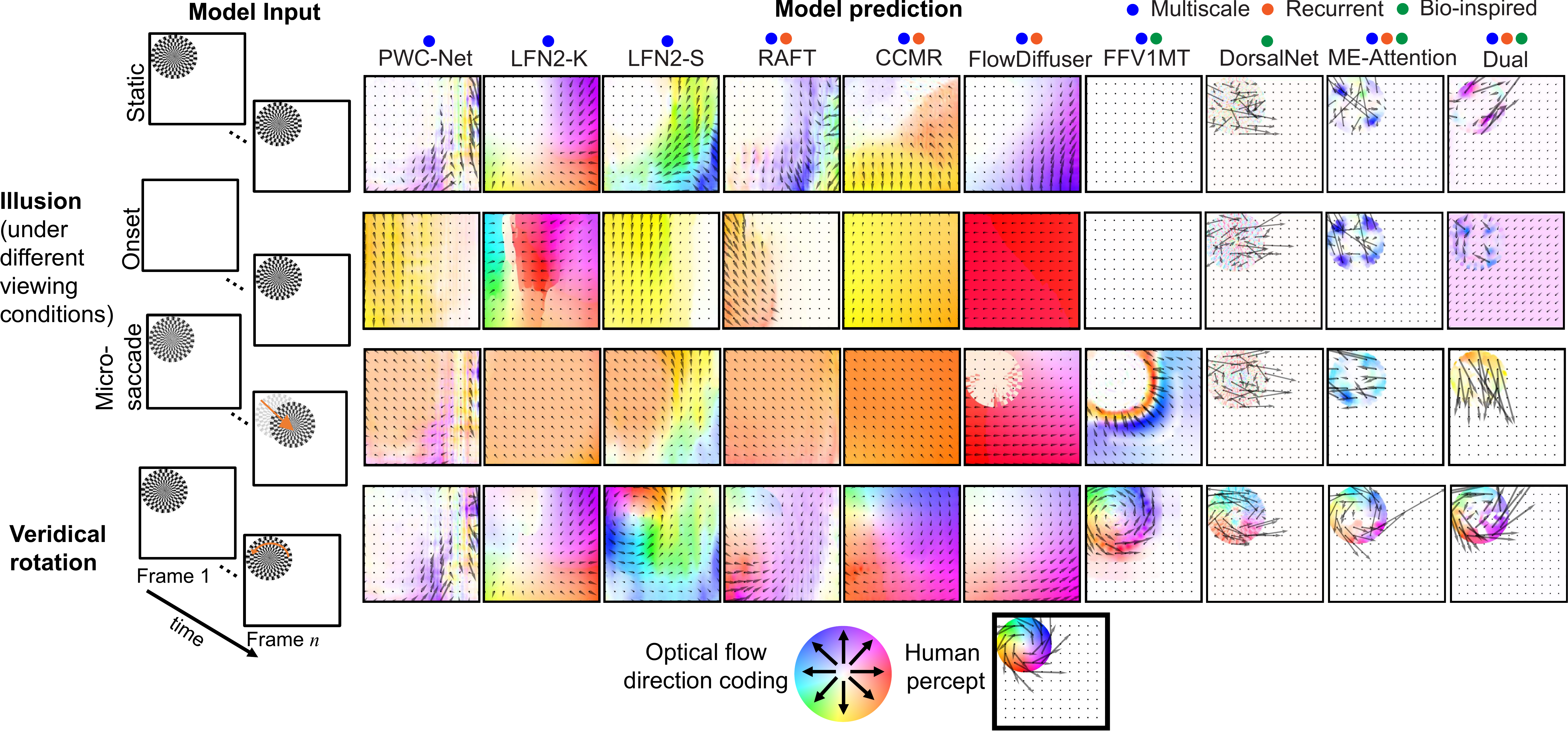}
  \caption{\textbf{Visualization of normalized model-predicted optical flow for grayscale stimuli under peripheral viewing condition.} 
Figure layout, color conventions, and direction-coding icons follow those introduced in Supplementary Figure~\ref{fig:supp_fig1_BY}. The ``Human percept'' icon (bottom right) illustrates the expected counterclockwise rotational flow perceived by human observers. The flow field was defined on the same uniform field and placed at the same spatial location as the stimulus, matching spatial extent and displacement offsets.}
  \label{fig:supp_fig2_G}
\end{figure*}

Peripheral viewing plays a central role in the Rotating Snakes illusion: human observers typically perceive robust illusory rotation when the stimulus lies in the visual periphery or when small eye movements produce transient retinal-image shifts. To emulate these conditions, we tested motion-estimation models on peripheral-viewing simulations in which the stimulus is embedded in a larger uniform field and translated across frames.

\vspace{-11px}
\paragraph{Implementation Details.}
 We generated sequences in which a $1386 \times 1386$ stimulus was embedded within a $2772 \times 2772$ uniform field. Two possible directions were used: bottom-right to top-left, or top-left to bottom-right. In each sequence, the image was shifted by a fixed displacement $\Delta$ per update, applied equally in the horizontal and vertical directions. We tested $\Delta \in \{15, 30, 60, 90, 120\}$ pixels, corresponding to the displacement magnitudes utilized in the central simulations described in the main text. 

\vspace{-11px}
\paragraph{Discussion.}
Under peripheral-viewing simulations (Fig.~\ref{fig:supp_fig2_G}), none of the models reproduce optical flow patterns consistent with the human percept. Only the bio-inspired architectures correctly identify the disk location and generate coherent rotational flow for the veridical-rotation condition. These results highlight a key distinction between engineering-oriented and neuroscience-inspired motion models. Engineering-oriented optical-flow networks fail under peripheral-viewing conditions because their computations rely on dense feature matching; large textureless regions provide few reliable keypoints, leading these models to produce spurious noise or to default to global smoothness priors. In contrast, models incorporating motion-energy mechanisms exhibit sparse, locally structured responses, reflecting activation of motion-energy units in textured regions and near-zero responses in uniform regions. Although DorsalNet does not include explicit motion-energy units, it contains units with tuning properties resembling biological motion-energy filters.

\newpage
\section{Effect of timing and direction of microsaccades and saccades on model perception}
\begin{figure}[ht]
  \centering
  \includegraphics[width=0.5\linewidth,height=0.5\textheight,keepaspectratio]{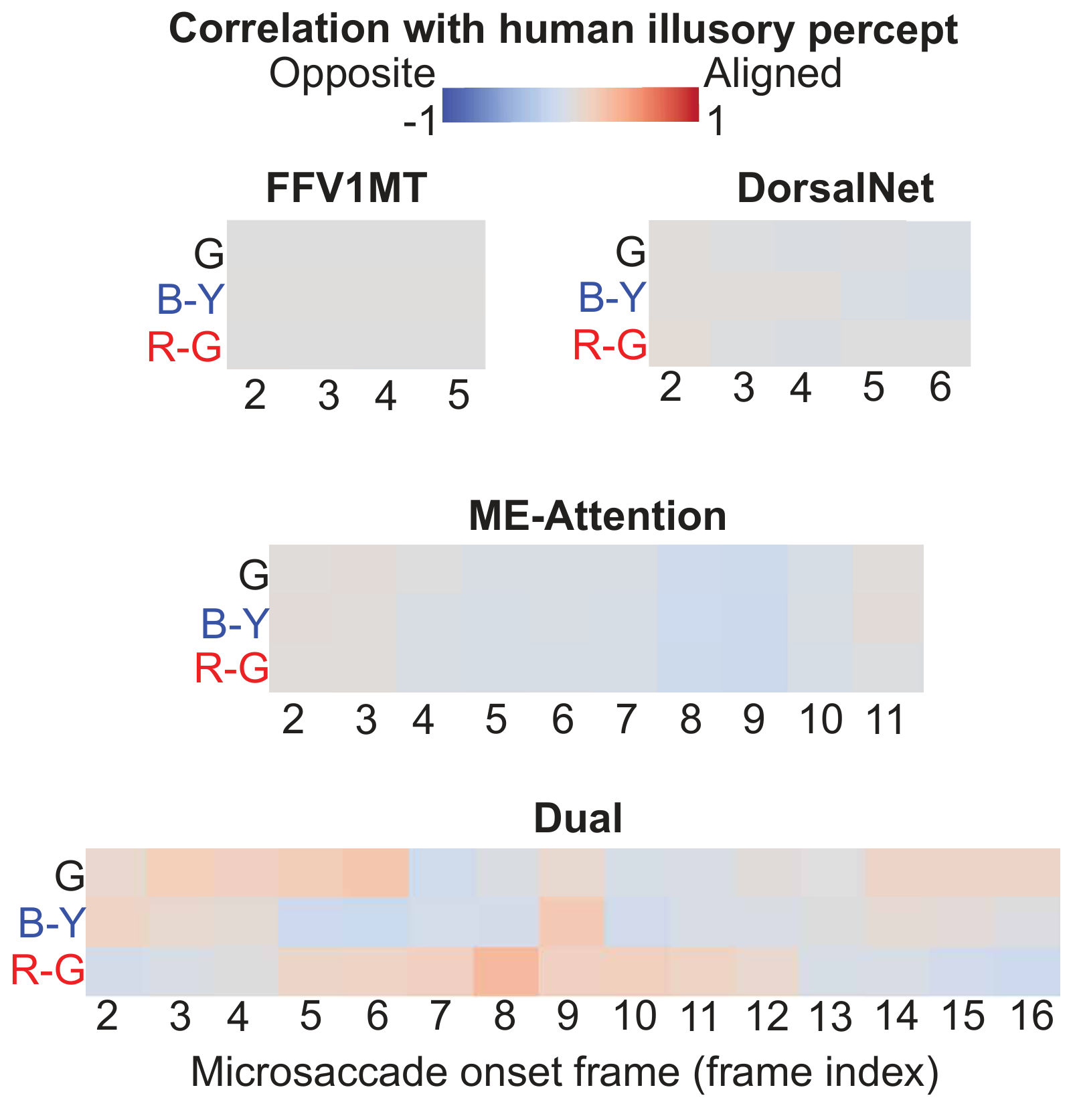}
  \caption{\textbf{Correlation across variations in microsaccade timing for bio-inspired models.} For each model, we simulated sequences in which a single microsaccadic displacement (30~px) occurred at different possible time points within the sequence (frame 2 through the final frame). Heatmaps show the correlation between model-predicited optical flow and the expected human illusory percept for grayscale (G), blue--yellow (B--Y), and red--green (R--G) variants.}
  \label{fig:supp_fig4_shift}
\end{figure}

\paragraph{Effects of microsaccade (and saccade) timing.} Because non–bio-inspired optical-flow models operate strictly on two-frame inputs, we examined timing effects only for multi-frame bio-inspired architectures. In this analysis, each sequence contained a \emph{single} 30-px microsaccadic displacement, introduced at different possible frames (from frame~2 to the final frame). Consistent with the main-text results, \textit{Dual} is the only model that shows clear positive alignment with human illusory perception (Figure ~\ref{fig:supp_fig4_shift}). Interestingly, for the red--green variant, Dual exhibits pockets of strong positive correlation when the shift occurs near the middle of the sequence, suggesting that the temporal position of the retinal slip can modulate the model's sensitivity to chromatic asymmetries in the stimulus. 

\begin{figure*}[t]
  \centering
  \includegraphics[width=\linewidth,height=0.72\textheight,keepaspectratio]{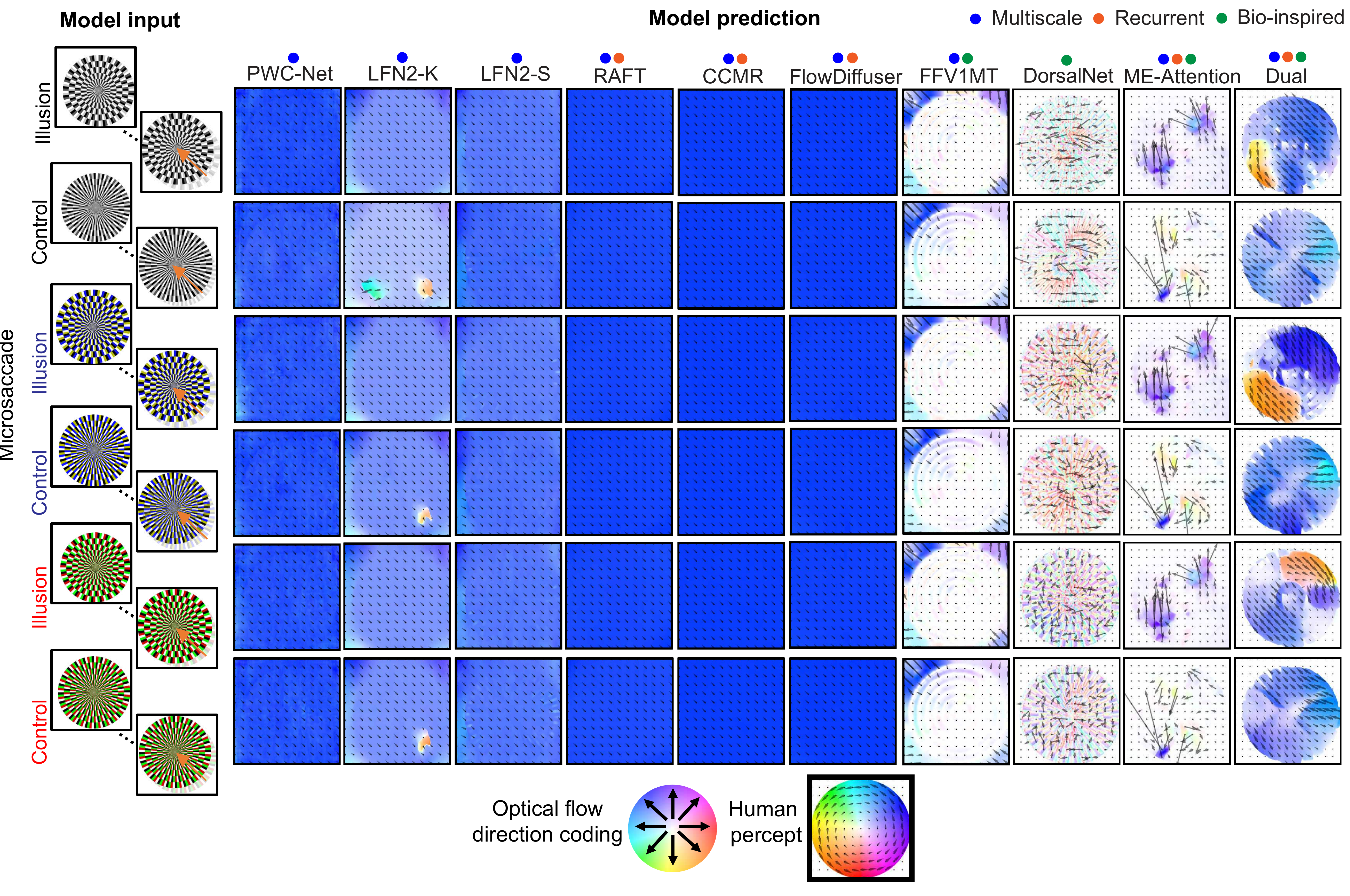}
  \caption{\textbf{Effect of reversing microsaccade direction on model-predicted optical flow.} 
We repeated the central-viewing microsaccade simulation from \cref{fig:fig2}, but applied the 
microsaccadic displacement in the \emph{opposite} direction. Model predictions are shown for illusion and control stimuli of three color variants across all models. The bottom icons follow those introduced in Supplementary Figure~\ref{fig:supp_fig1_BY}.}
  \label{fig:supp_fig5_fig_opp}
\end{figure*}

\begin{figure}[b]
  \centering
  \includegraphics[width=0.5\linewidth,height=0.5\textheight,keepaspectratio]{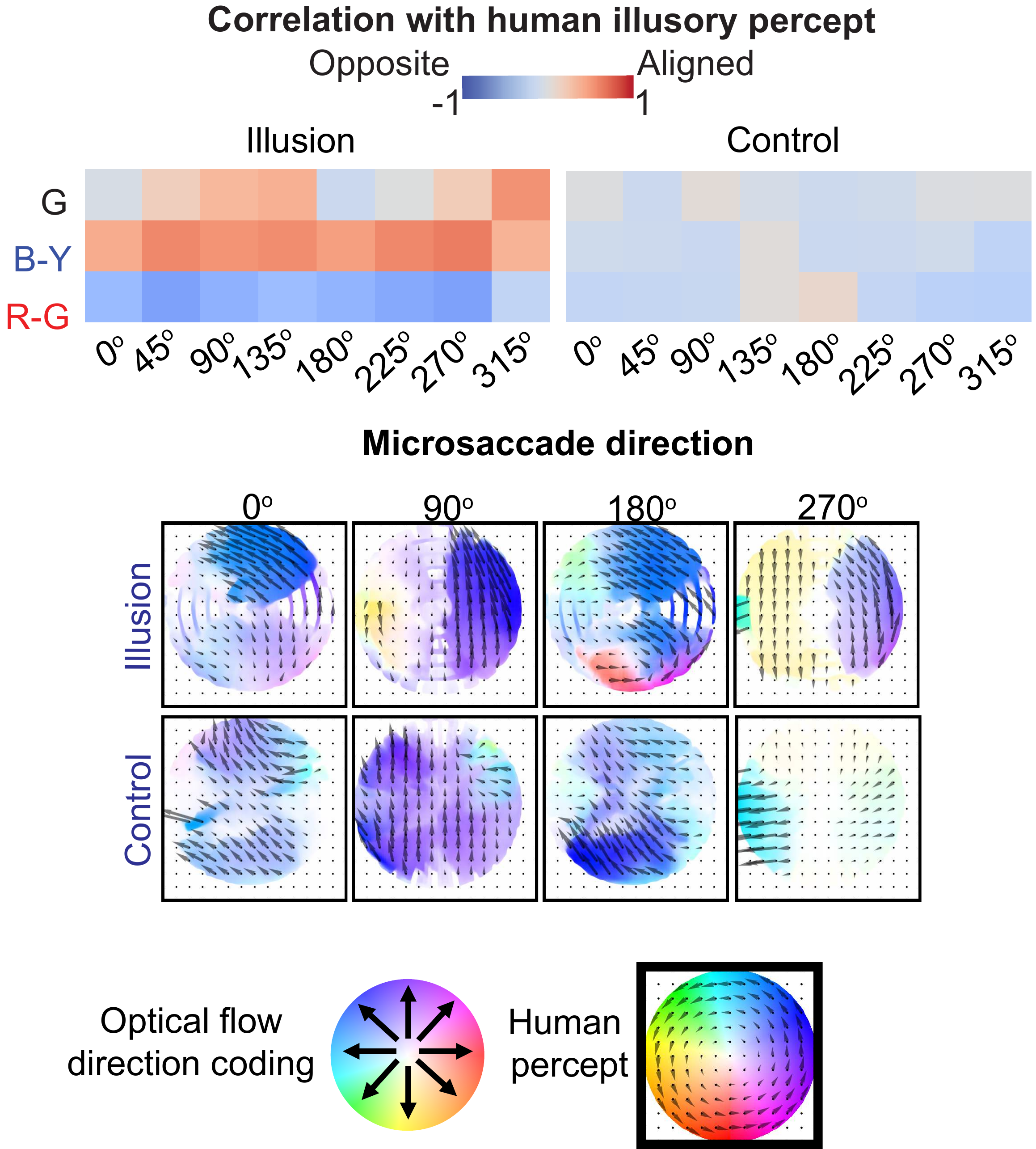}
  \caption{\textbf{Effect of microsaccade direction on Dual model predictions.}
 Correlations are shown for illusion (left) and control (right) stimuli across eight possible microsaccade directions. The direction of the microsaccadic displacement was varied while keeping all other parameters identical to those used in \cref{fig:fig2} and \cref{fig:supp_fig5_fig_opp}. Qualitative flow fields for the blue--yellow variant (bottom) illustrate how Dual's predicted rotation varies with displacement direction. The bottom icons follow those introduced in Supplementary Figure~\ref{fig:supp_fig1_BY}.}
  \label{fig:supp_fig6_multi_dir_BY}
\end{figure}

\vspace{-11px}
\paragraph{Effects of microsaccade (and saccade) direction.}
Reversing the microsaccade direction yields qualitative trends consistent with those observed in \cref{fig:fig2}, with model behavior largely invariant to the shift polarity (Figure~\ref{fig:supp_fig5_fig_opp}). As in the main analysis, engineering-oriented models (PWC-Net, LFN2-S, RAFT, CCMR, FlowDiffuser) produce near-uniform or noisy flows dominated by the imposed displacement, and show no evidence of rotation-like organization for either illusion or control stimuli. Only Dual shows partially coherent rotation-like flow for 
the illusion stimulus and results for control stimuli remain near-zero or inconsistent.

We additionally varied the direction of microsaccadic displacements (Figure~\ref{fig:supp_fig6_multi_dir_BY}) while holding all other simulation parameters identical to those used in \cref{fig:fig2}. Eight displacement orientations were tested: $0^{\circ}$, $45^{\circ}$, \ldots, $315^{\circ}$, with a fixed magnitude of 30~px. Overall, the results are consistent with those reported in the main text: positive correlations for the blue--yellow and negative correlations for the red--green variants. Although the model consistently encodes counterclockwise flow for the blue--yellow illusion variant, the \emph{spatial location} at which this rotational pattern emerges depends strongly on the direction of retinal displacement (Figure~\ref{fig:supp_fig6_multi_dir_BY}, bottom). These results demonstrate that the Dual model produces illusion effect only in regions where the microsaccade produces sufficiently large local luminance changes along its first- and second-order motion pathways. Because the Rotating Snakes pattern contains directionally asymmetric luminance ramps, different microsaccade directions emphasize different subsets of local units, leading to spatially shifted “islands” of counterclockwise flow. Importantly, control stimuli fail to generate such structured rotational patterns, indicating that the observed responses depend on the luminance asymmetries that drive the illusion rather than on the displacement trajectory alone.

\begin{figure}[ht]
  \centering
  \includegraphics[width=0.5\linewidth,height=0.5\textheight,keepaspectratio]{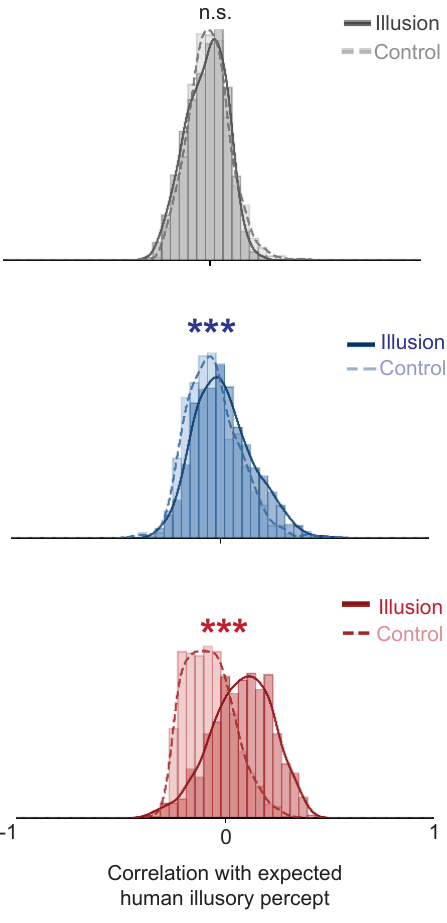}
  \caption{\textbf{Distribution of human-alignment scores of Dual model responses under random microsaccades or saccades.} Histograms show the distribution of correlations between predicted optical
flow and the expected human illusory percept for grayscale (top), blue--yellow (middle), and red--green (bottom) variants across 1000 simulations. Asterisks denote significant differences between illusion and control distributions (One-sided Wilcoxon signed rank test).}
  \label{fig:supp_fig_8_correlation_dist}
\end{figure}

\vspace{-11px}
\paragraph{Randomized microsaccade (and saccade) simulations.}
\label{para:random_saccades}
We relaxed the experimenter-controlled constraints used in the main text and evaluated the
Dual model under more naturalistic retinal-slip statistics (Figure~\ref{fig:supp_fig_8_correlation_dist}). Each sequence contained up to
three microsaccades with (i) a random onset frame sampled uniformly across the full
duration, (ii) a random direction drawn from eight possible orientations
$\{0^{\circ}, 45^{\circ}, \ldots, 315^{\circ}\}$, and (iii) a random displacement magnitude drawn from $\{15, 30, 60, 90, 120\}$~px. We generated 1000 independent simulations for each color variant (grayscale, blue--yellow, red--green). Model--human alignment was computed as in the main analysis, and statistical significance was assessed using \emph{one-sided Wilcoxon signed-rank} tests evaluating (a) whether illusion correlations were significantly greater
than zero, (b) whether control correlations were significantly greater than zero, and
(c) whether illusion correlations were significantly greater than control correlations.

Across randomized simulations, the grayscale illusion condition yielded correlations that
were significantly less than zero ($p=3.14 \times 10^{-8}$), and control
correlations were non-significant ($p>0.05$). Consistent with this,
illusion and control distributions did not differ ($p>0.05$). For the blue--yellow variant, illusion correlations were significantly greater than zero ($p=4.75 \times 10^{-4}$), while control correlations were significantly less than zero ($p=2.38 \times 10^{-15}$). Illusion correlations were significantly larger than control correlations ($p=1.21 \times 10^{-15}$), indicating robust human-aligned rotational structure under naturalistic retinal slip.

The red--green variant showed illusion correlations significantly greater than zero ($p=8.47 \times 10^{-57}$), while control correlations were significantly less than zero ($p=1.09 \times 10^{-79}$). Illusion correlations were significantly larger than control correlations ($p=6.97 \times 10^{-120}$). Notably, although controlled-direction microsaccades in ~\cref{fig:fig2,fig:supp_fig6_multi_dir_BY} often produced correlations opposite to the human percept for red--green stimuli, randomized microsaccades yielded predominantly positive alignment.

\clearpage
\section{Other evaluation metrics}
For completeness, we report two standard optical flow evaluation metrics: average endpoint error and average angular error \cite{baker2011database}. Let $\mathbf{P} \in \mathbb{R}^{n \times 2}$ denote the predicted flow field and $\mathbf{R} \in \mathbb{R}^{n \times 2}$ denote the ground-truth (veridical) flow field at $n$ spatial locations. Each row $\mathbf{P}_{i,:} = (u_i, v_i)$ and $\mathbf{R}_{i,:} = (u_i^{*}, v_i^{*})$ corresponds to the horizontal and vertical components of the flow at spatial location $i$.

\vspace{-11pt}
\paragraph{Average Endpoint Error (EPE).} 
The endpoint error measures the Euclidean distance between predicted and ground-truth flow vectors:
\begin{equation}
\text{EPE}_i = \|\mathbf{P}_{i} - \mathbf{R}_{i}\|_2,
\label{eq:epe}
\end{equation}
where $\|\cdot\|_2$ denotes the $\ell_2$ norm in $\mathbb{R}^2$. We report the mean EPE averaged over all spatial locations:
\[
\overline{\text{EPE}} = \frac{1}{n}\sum_{i=1}^{n}\text{EPE}_i.
\]

\vspace{-11pt}
\paragraph{Average Angular Error (AE).}
Following standard practice, we embed each 2D flow vector $(u,v)$ into 3D space as $(1,u,v)$ to avoid degeneracies when vectors have small magnitude, and we compute the angular deviation between predicted and ground-truth flow as:
\begin{equation}
\text{AE}_{i}
= \arccos
\left(
    \frac{
        1 + u_{i} u^{*}_{i} + v_{i} v^{*}_{i}
    }{
        \sqrt{1 + u_{i}^{2} + v_{i}^{2}}
        \;\sqrt{1 + {u^{*}_{i}}^{2} + {v^{*}_{i}}^{2}}
    }
\right).
\label{eq:ae}
\end{equation}
We report the mean angular error averaged over all spatial locations:
\[
\overline{\text{AE}} = \frac{1}{n}\sum_{i=1}^{n}\text{AE}_i.
\]

Lower AAE indicates better directional agreement between predicted and
veridical flow fields.

\begin{figure*}[t]
  \centering
  \includegraphics[width=0.6\linewidth,height=0.6\textheight,keepaspectratio]{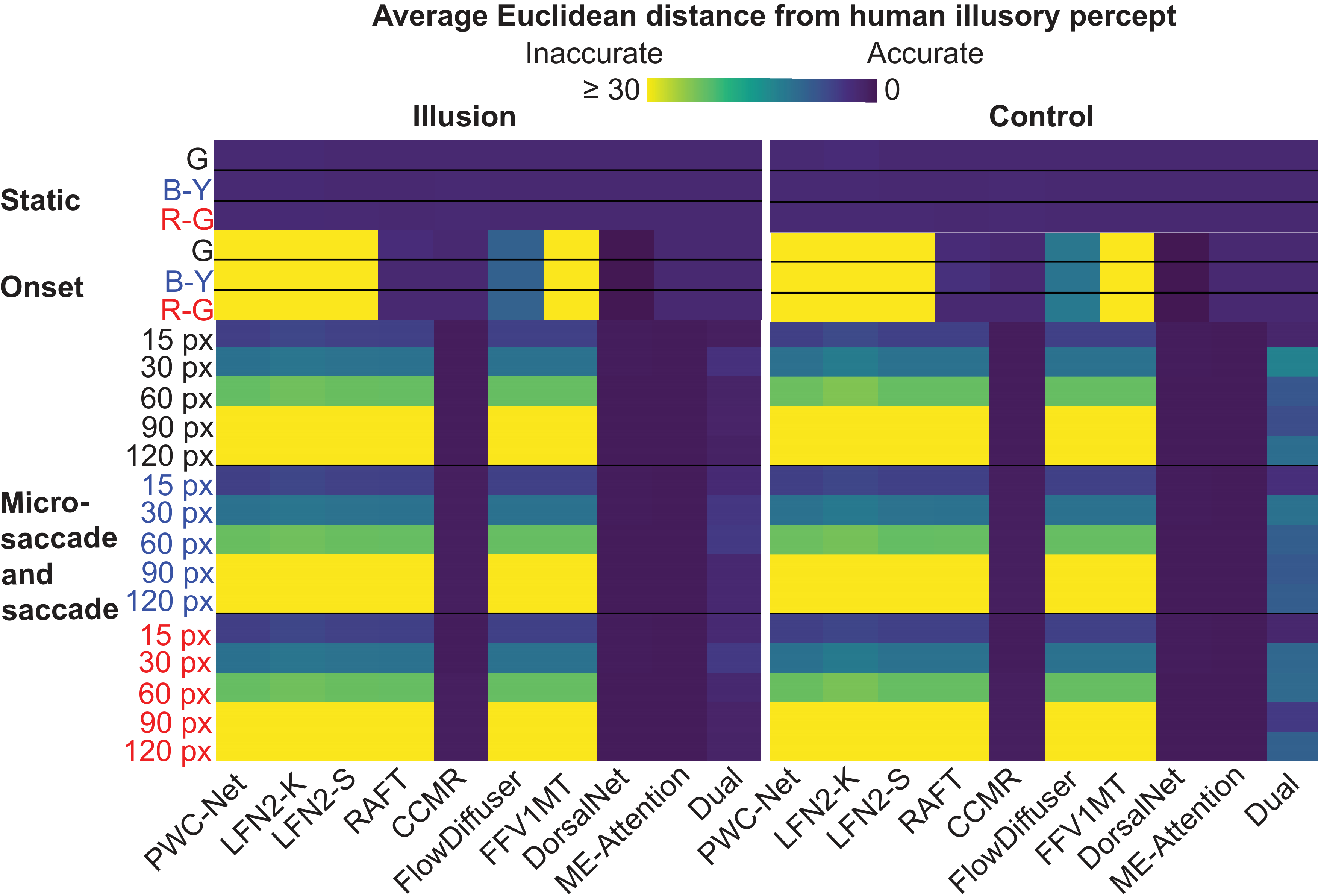}
  \caption{\textbf{Average endpoint error (EPE) across models and stimuli.}
Heatmaps show the mean Euclidean distance between the model-predicted flow
and the human illusory-percept flow under each viewing condition
(static, onset, and microsaccade/saccade) for illusion (left) and control (right)
stimuli (same simulations as ~\cref{fig:fig3}). }
  \label{fig:supp_fig3_EPE}
\end{figure*}

\begin{figure*}[t]
  \centering
  \includegraphics[width=0.6\linewidth,height=0.6\textheight,keepaspectratio]{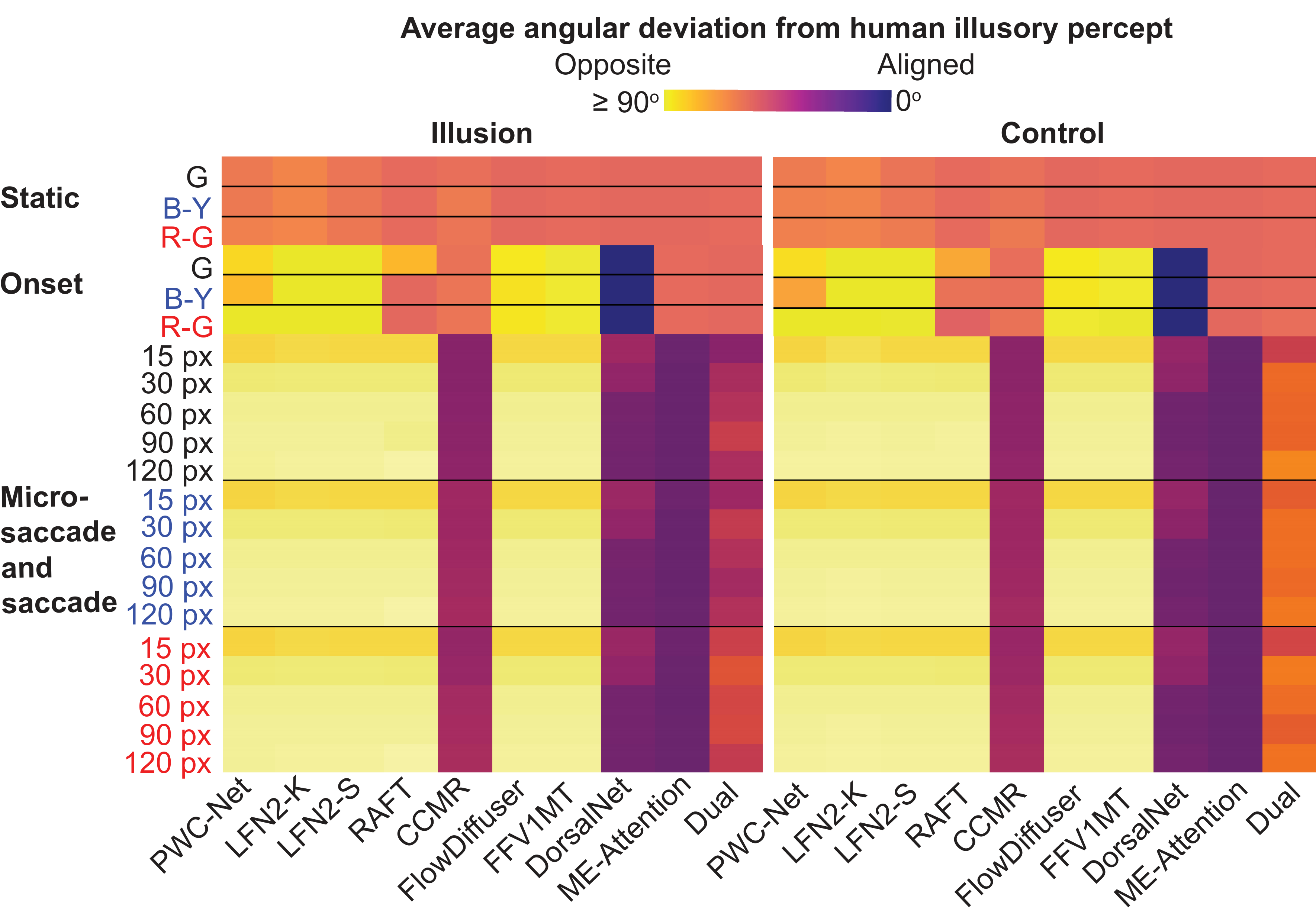}
  \caption{\textbf{Average angular error (AE) across models and stimuli.} The simulations and layout follow those introduced in \ref{fig:supp_fig3_EPE}.  
} 
  \label{fig:supp_fig3_AE}
\end{figure*}

\vspace{-11px}
\paragraph{Interpretation of alternative metrics.} The endpoint error (EPE) and angular error (AE) capture different properties of the flow fields than the correlation metric used in the main text (Figs. ~\ref{fig:supp_fig3_EPE} and ~\ref{fig:supp_fig3_AE}). Whereas correlation measures \emph{global directional and magnitude consistency} with the canonical counterclockwise rotational field, EPE and AE evaluate \emph{local vector-wise agreement in magnitude} (EPE) or \emph{directional deviation at each spatial location} (AE), independent of global structure.

These distinctions explain why models such as \emph{DorsalNet} appear to achieve
high alignment under EPE and AE despite showing little coherent rotation in
Fig. ~\ref{fig:fig3}. As seen in Figs. ~\ref{fig:supp_fig1_BY} and ~\ref{fig:supp_fig1_RG}, DorsalNet often produces flow fields with several localized regions of strong motion energy, including patches whose directions partially resemble rotational flow. Because EPE and AE are averaged over spatial locations, these high-magnitude local vectors can reduce the mean error—even when the global pattern does not form a coherent rotation and even when similar rotational components also appear in the control condition.

In contrast, the correlation metric emphasizes whether the overall spatial
organization of the flow agrees with the human percept. It penalizes cases
where models produce scattered local vectors or noisy fields. Under this more stringent criterion, only Dual captures a globally consistent counterclockwise structure, whereas models like DorsalNet display poorly organized flow that inflates performance under EPE and AE but fails to align perceptually.

\newpage
\section{Probing, control, and ablation analysis}
\begin{figure}[h]
  \centering
  \includegraphics[width=0.5\linewidth,height=0.5\textheight,keepaspectratio]{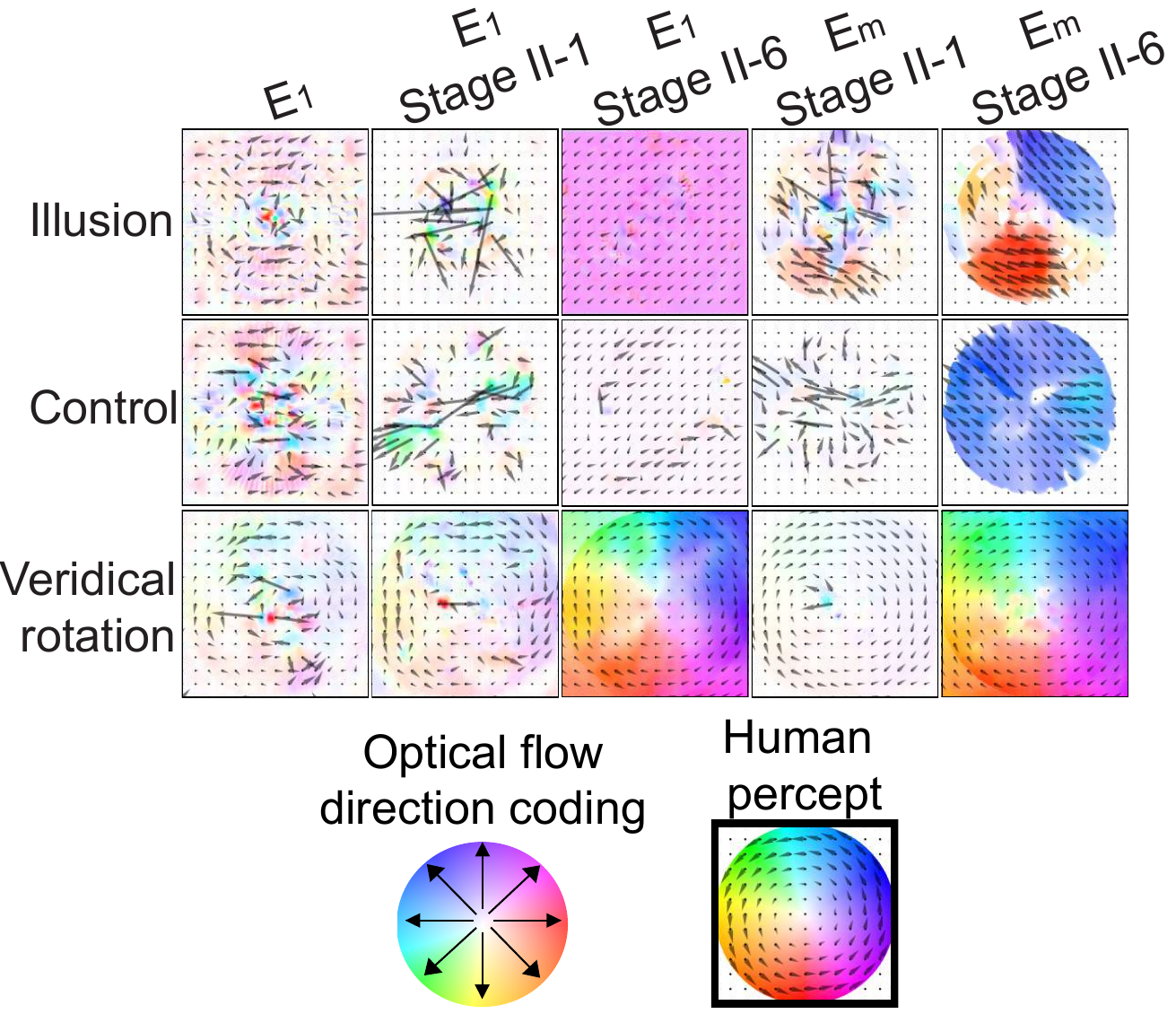}
  \caption{\textbf{Visualization of decoded flow fields from representations at different processing stages.} Shown are the first-order pathway alone ($E_{1}$), the first and last recurrent iterations applied to $E_{1}$ (Stage~II-1 and Stage~II-6), and the corresponding first and last recurrent iterations applied to $E_m$.}
  \label{fig:supp_fig_pathway}
\end{figure}

\paragraph{Probing analysis.}
 Fig.~\ref{fig:supp_fig_pathway} shows qualitative examples of optical flow across pathways and recurrent stages. Flow fields decoded from $E_1$ alone primarily capture local flicker-like motion and lack any coherent tangential structure for the illusion. The addition of the higher-order channel visibly alters the predicted flow fields. The fused representation $E_m$ (Stage~I) produces stronger, spatially organized flow for illusion stimuli. This improvement highlights the role of higher-order motion features in generating human-like percepts.

Recurrent integration further amplifies and stabilizes rotational motion. Final-stage outputs (Stage~II-6) exhibit globally counterclockwise flow for the illusion condition, closely resembling the veridical rotational field, despite remaining spatially nonuniform. In contrast, control stimuli do not show increases in alignment across iterations, confirming that recurrence selectively enhances features specific to the illusory image rather than simply amplifying input transients.

\begin{figure}[h]
  \centering
  \includegraphics[width=\linewidth,height=0.75\textheight,keepaspectratio]{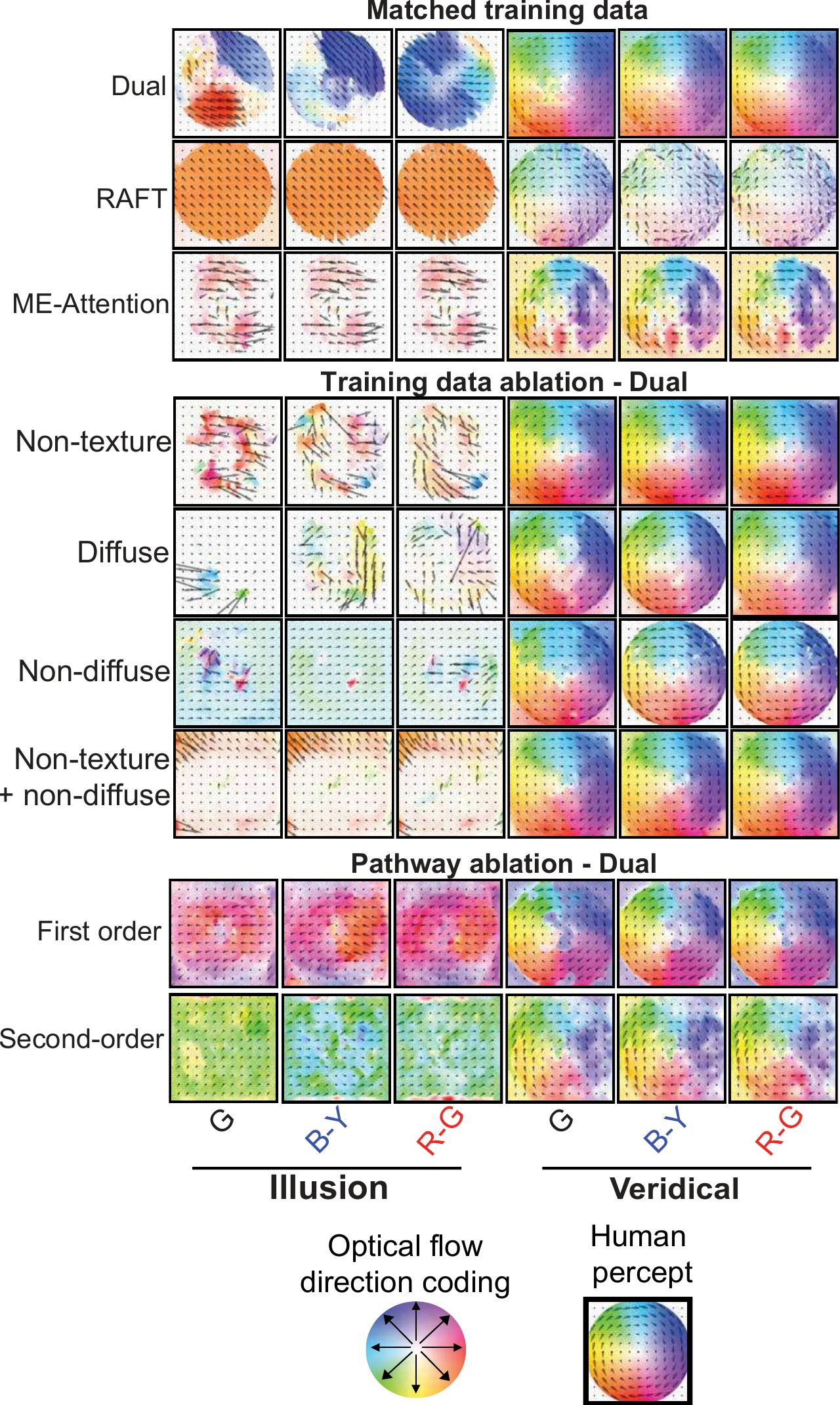}
  \caption{\textbf{Predicted flow fields of retrained models evaluated on illusion and veridical-rotation stimuli.} Shown are Dual, RAFT, and ME-Attention trained on the full dataset, as well as Dual trained with data and architectural ablations.}
    \label{fig:supp_fig_ablation}
\end{figure}

\paragraph{Control Analysis}
We retrained RAFT and ME-Attention on the same training data as Dual. Fig.~\ref{fig:supp_fig_ablation} (top panel) shows qualitative optical flow examples under the 30-px microsaccade condition and veridical rotation. RAFT primarily captures the shift direction of the microsaccade, whereas ME-Attention shows some sensitivity to local luminance structure but lacks global coherence, with regions predicting rightward motion. For veridical rotation, both models capture the expected motion, although with reduced global coherence and weaker speed than models trained on their original training data (see Fig.~\ref{fig:fig2} in the main text).

\paragraph{Ablation analysis}
Qualitative examples of model-predicted flow after retraining Dual with subsets of the training data and ablating first- and second-order pathways are shown in Fig.~\ref{fig:supp_fig_ablation} (middle and bottom panels, respectively); all variants produced accurate optical flow for veridical rotation. Removing the non-texture object motion training subset leads to noisy local flow inconsistent with global rotation. Removing the diffuse dataset leads to large regions of near-zero flow, with regions of local flow also inconsistent with rotational motion. In contrast, removing the non-diffuse dataset produces predominantly globally coherent but weakly predicted flow, inconsistent with both microsaccade direction and rotational motion. Removing both non-texture and non-diffuse subsets results in flow predictions consistent with the microsaccade direction rather than rotation. Removing the first-order pathway largely eliminates sensitivity to local luminance structure; predicted optical flows are dominated by weak, rightward flow. Removing the second-order pathway produces near-uniform flow with pockets predicting flow in an opposing direction. Predictions for veridical rotation after ablating the first-order pathway align with true rotation, while ablating the second-order pathway introduces regions of weak flow.

\newpage
\section{Other anomalous-motion illusions}
We have shown results under random microsaccade simulations in Fig.~\ref{fig:fig6} (see implementation details in Appendix 4). Here, figues~\ref{fig:supp_fig_7_peripheral1} and~\ref{fig:supp_fig_7_ouchi} show results under the same simulated viewing conditions used in Fig.~\ref{fig:fig2} (static, onset, and microsaccades).

\vspace{-11px}
\paragraph{Peripheral and central drift illusions.}
All three stimuli (two peripheral drift variants and the central drift illusion) elicit a robust clockwise percept for human observers. Under both the static and onset conditions, none of the evaluated models produced flow patterns resembling this perceptual experience. Consistent with the results in Fig.~\ref{fig:fig6}, only the Dual model exhibited any degree of coherent rotational structure under simulated microsaccades, and this effect emerged solely for the peripheral drift illusions. For the blue--yellow stimulus (middle column), Dual generated a clear clockwise pattern that aligned with the human percept. In contrast, for the grayscale variant (left column), Dual produced a predominantly \emph{counterclockwise} pattern, opposite to the expected direction. The central drift illusion did not induce meaningful rotation in any model.

\vspace{-11px}
\paragraph{Ouchi illusion.}
ME-Attention and Dual exhibit figure–ground segmentation: both models generated distinct motion responses for the central region and its surround even under static and onset presentations. However, the segmentation boundaries were sometimes weak or incomplete—for example, Dual showed degraded segmentation for both the onset and microsaccade conditions. Critically, although ME-Attention and Dual captured coarse region-wise separation, the direction of the predicted flow may not align with the illusory percept experienced by human observers.

\begin{figure*}[tp]
  \centering
  \includegraphics[width=\linewidth,height=0.72\textheight,keepaspectratio]{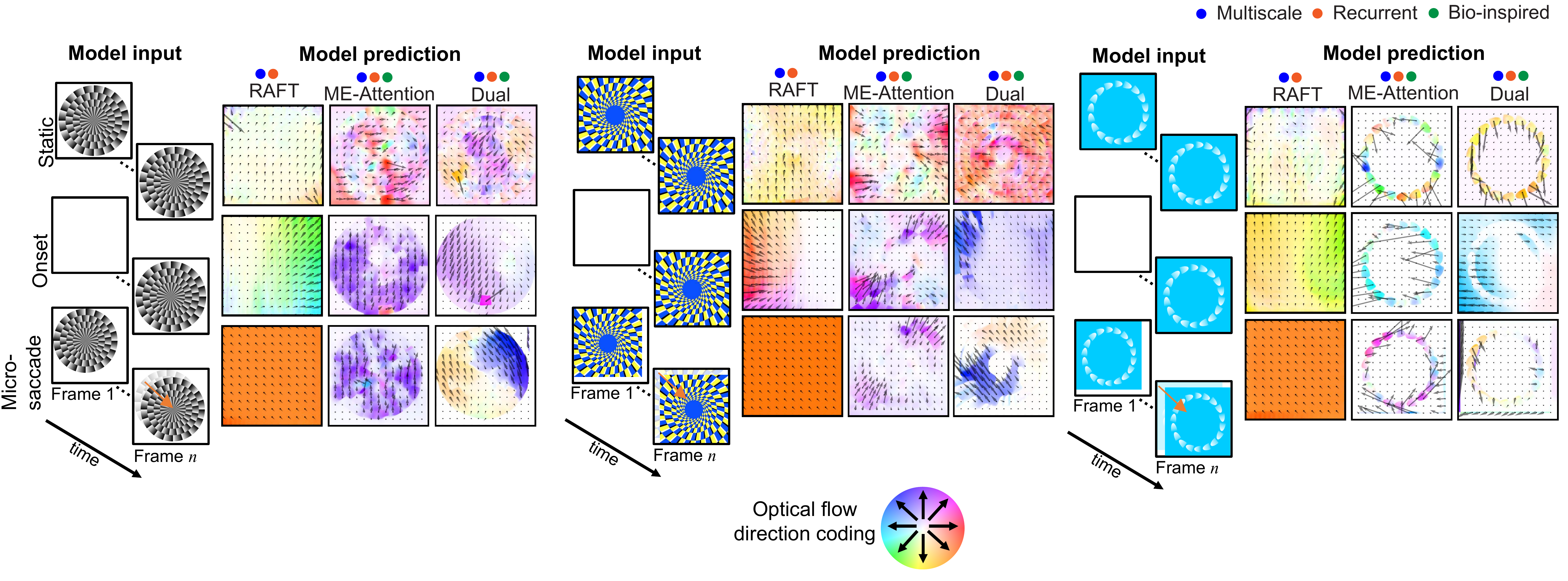}
  \caption{\textbf{Visualization of normalized model-predicted optical flow for peripheral drift illusions (PDI) and the central drift illusion (CDI) across selected models.} The layout follows the conventions introduced in ~\cref{fig:supp_fig1_BY}. This figure extends the results shown in Figure~6 by presenting the corresponding PDI variants and the CDI under the same subset of simulated conditions as~\cref{fig:supp_fig1_BY}. 
} 
  \label{fig:supp_fig_7_peripheral1}
\end{figure*}

\begin{figure*}[tp]
  \centering
  \includegraphics[width=\linewidth,height=0.72\textheight,keepaspectratio]{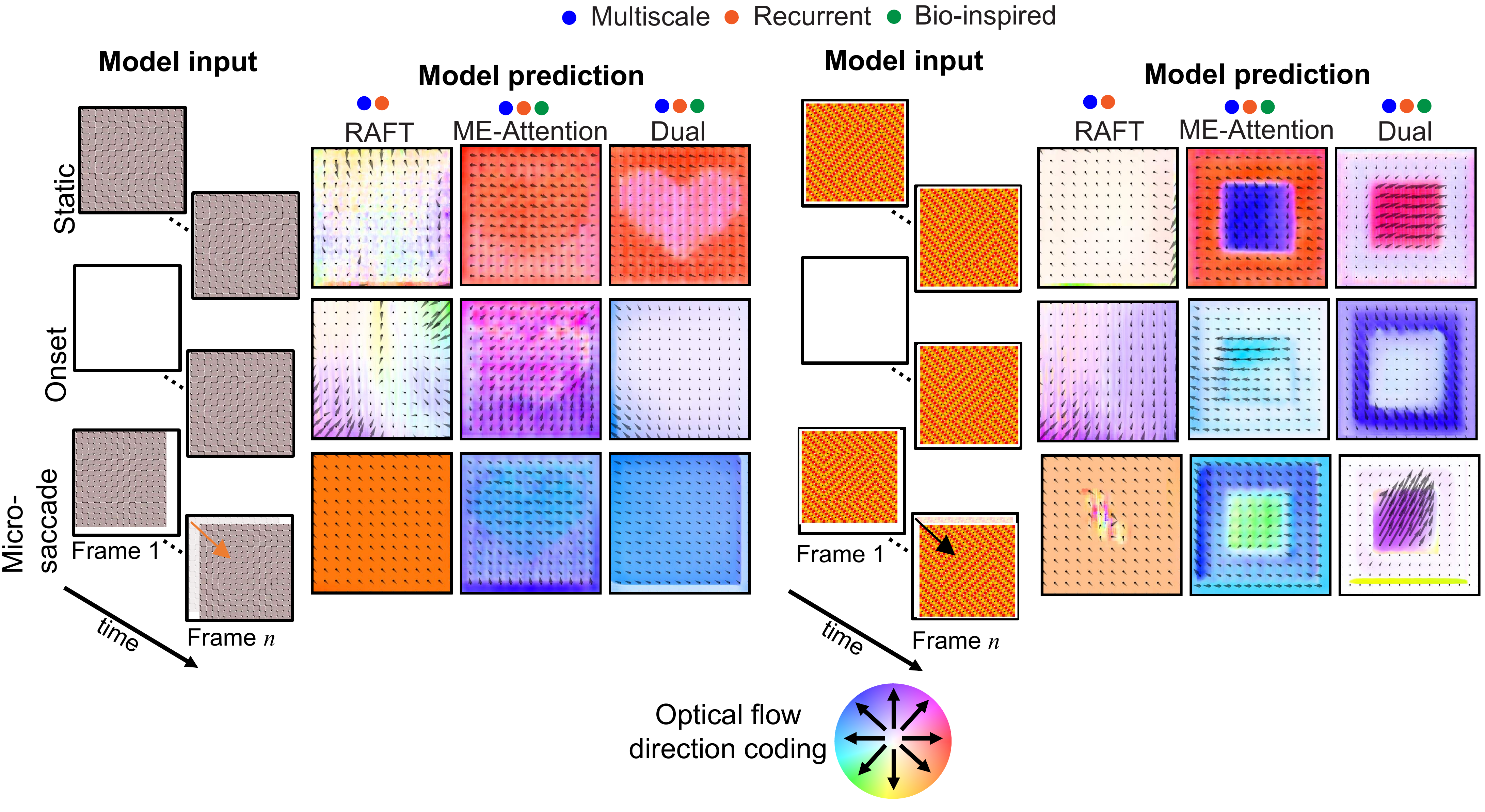}
  \caption{\textbf{Visualization of normalized model-predicted optical flow for Ouchi illusions.} 
The layout follows the conventions introduced in~\cref{fig:supp_fig1_BY}. 
This figure complements the results in Figure~6 by showing the corresponding Ouchi variants evaluated under the same subset of simulated conditions as~\cref{fig:supp_fig1_BY}.
}
  \label{fig:supp_fig_7_ouchi}
\end{figure*}

\newpage
\section{Movie demonstration}
We include video demonstrations illustrating three representative examples of the simulated viewing conditions used as model inputs. All videos are rendered at 5 Hz to aid visualization. Each video includes a fixation cross placed in the bottom-right corner to facilitate stable fixation.

\paragraph{Demonstration of stimulus-onset simulation.}
Video 1 demonstrates the onset simulation for the grayscale illusion stimulus.

\paragraph{Demonstrations of microsaccade and saccade simulations.}
Videos 2 and 3 illustrate the single-shift simulations used for microsaccade and saccade conditions. Video~2 presents a 30~px shift magnitude for the grayscale illusion stimulus, whereas Video~3 shows a 120~px shift magnitude for the blue--yellow illusion stimulus. 

\end{document}